\documentclass{article}

\usepackage{microtype}
\usepackage{booktabs} % for professional tables

\usepackage{hyperref}

\usepackage[accepted]{icml2024}

\usepackage{amsmath}
\usepackage{amssymb}
\usepackage{mathtools}
\usepackage{amsthm}

\usepackage[pdftex]{graphicx}
\usepackage{multirow}
\usepackage{listings}

\usepackage{caption}
\usepackage[subrefformat=parens]{subcaption}
\usepackage{comment}
\usepackage{ulem}

\usepackage[capitalize,noabbrev]{cleveref}

\theoremstyle{plain}

\theoremstyle{definition}

\theoremstyle{remark}

\usepackage[textsize=tiny]{todonotes}

\icmltitlerunning{ResBit: Residual Bit Vectors for Categorical Values}

\begin{document}

\twocolumn[
\icmltitle{ResBit: Residual Bit Vectors for Categorical Values}



\icmlsetsymbol{equal}{*}

\begin{icmlauthorlist}
\icmlauthor{Masane Fuchi}{yyy}
\icmlauthor{Amar Zanashir}{comp}
\icmlauthor{Hiroto Minami}{comp}
\icmlauthor{Tomohiro Takagi}{yyy}
% \icmlauthor{Firstname2 Lastname2}{equal,yyy,comp}
% \icmlauthor{Firstname3 Lastname3}{comp}
\end{icmlauthorlist}

\icmlaffiliation{yyy}{Department of Computer Science, Meiji University, Kanagawa, Japan}
\icmlaffiliation{comp}{LAC Co. Ltd., Tokyo, Japan}

\icmlcorrespondingauthor{Masane Fuchi}{ce235031@meiji.ac.jp}

\icmlkeywords{Machine learning, tabular data generation, curse of dimensionality}
\vskip 0.3in
]



\printAffiliationsAndNotice{}  % leave blank if no need to mention equal contribution

\begin{abstract}
One-hot vectors, a common method for representing discrete/categorical data, in machine learning are widely used because of their simplicity and intuitiveness. However, one-hot vectors suffer from a linear increase in dimensionality, posing computational and memory challenges, especially when dealing with datasets containing numerous categories. In this paper, we focus on tabular data generation, and reveal the multinomial diffusion faces the mode collapse phenomenon when the cardinality is high. Moreover, due to the limitations of one-hot vectors, the training phase takes time longer in such a situation. To address these issues, we propose \textbf{Res}idual \textbf{Bit} Vectors (ResBit), a technique for densely representing categorical data. ResBit is an extension of analog bits and overcomes limitations of analog bits when applied to tabular data generation. Our experiments demonstrate that ResBit not only accelerates training but also maintains performance when compared with the situations before applying ResBit. Furthermore, our results indicate that many existing methods struggle with high-cardinality data, underscoring the need for lower-dimensional representations, such as ResBit and latent vectors.
\end{abstract}

\section{Introduction}

\begin{figure*}[htbp]
  \vskip 0.1in
  \centering
  \includegraphics[width=0.8\linewidth]{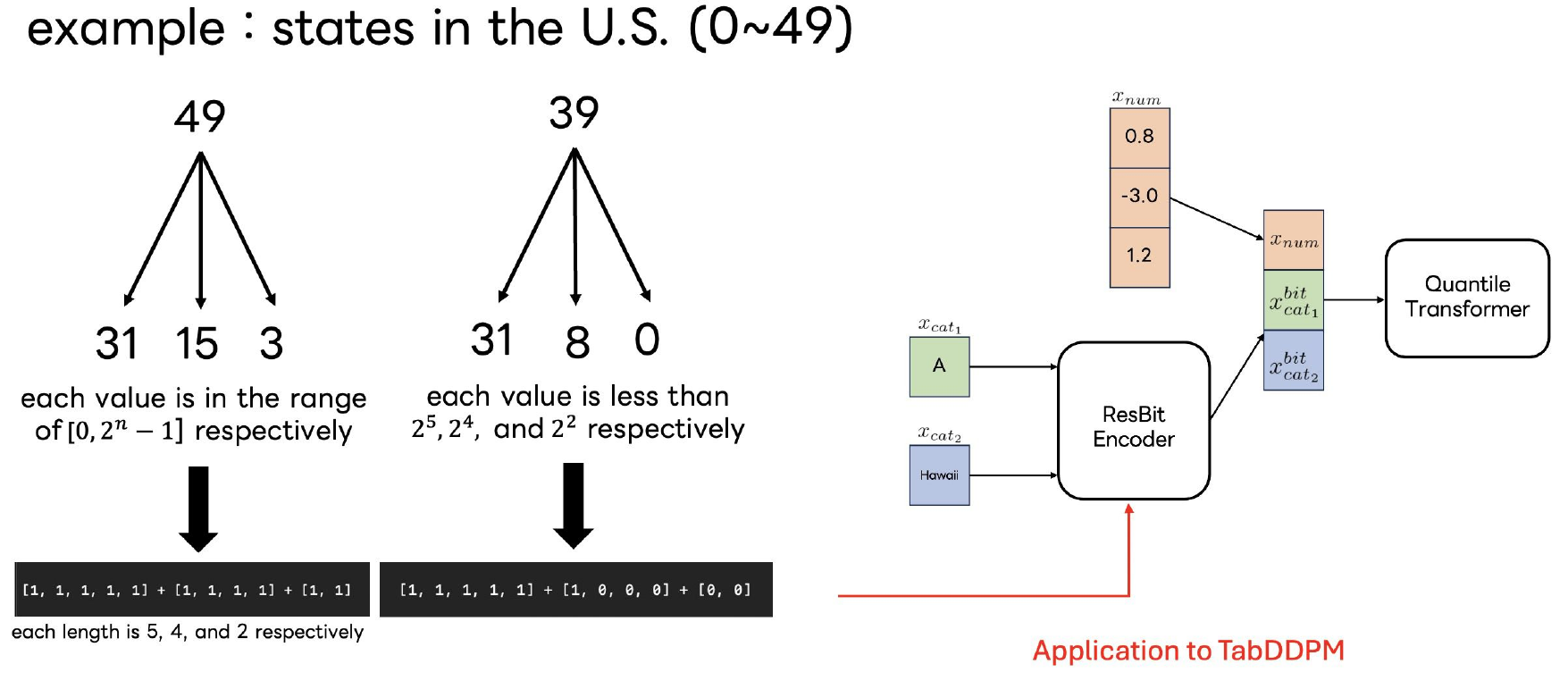}
  \caption{Overview of the proposed ResBit (left) and application example to TabDDPM (right).}
  \label{fig:resbit}
  \vskip 0.1in
\end{figure*}

The application of machine learning in real-world scenarios is increasing, and one notable case is the use of tabular data. Tabular data consists of complex information, encompassing numerical data representing continuous values and categorical data representing discrete quantities. Consequently, decision tree-based methods without using deep learning still dominate tabular data analysis due to its intricate nature~\cite{grinsztajn2022why}. Handling categorical data in machine learning involves converting it into numerical form, and deep learning commonly employs one-hot encoding for this purpose. Examples of this encoding technique extend beyond tabular data analysis and are observed in other domains, such as label or class assignment in image classification and class-conditional image generation.

One-hot vectors are widely utilized due to their simplicity and ease of implementation. However, they come with drawbacks such as high memory consumption due to sparsity and an increase in computational complexity as dimensions grow. When considering the representation of categorical data using one-hot vectors, the challenges posed by the ``curse of dimensionality'' make it difficult to handle in machine learning. The issue arises from the increased dimensionality leading to increased memory usage and computational demands, making the practical implementation in machine learning challenging.

\begin{figure}[htbp]
  \vskip 0.1in
  \centering
  \includegraphics[width=0.7\linewidth]{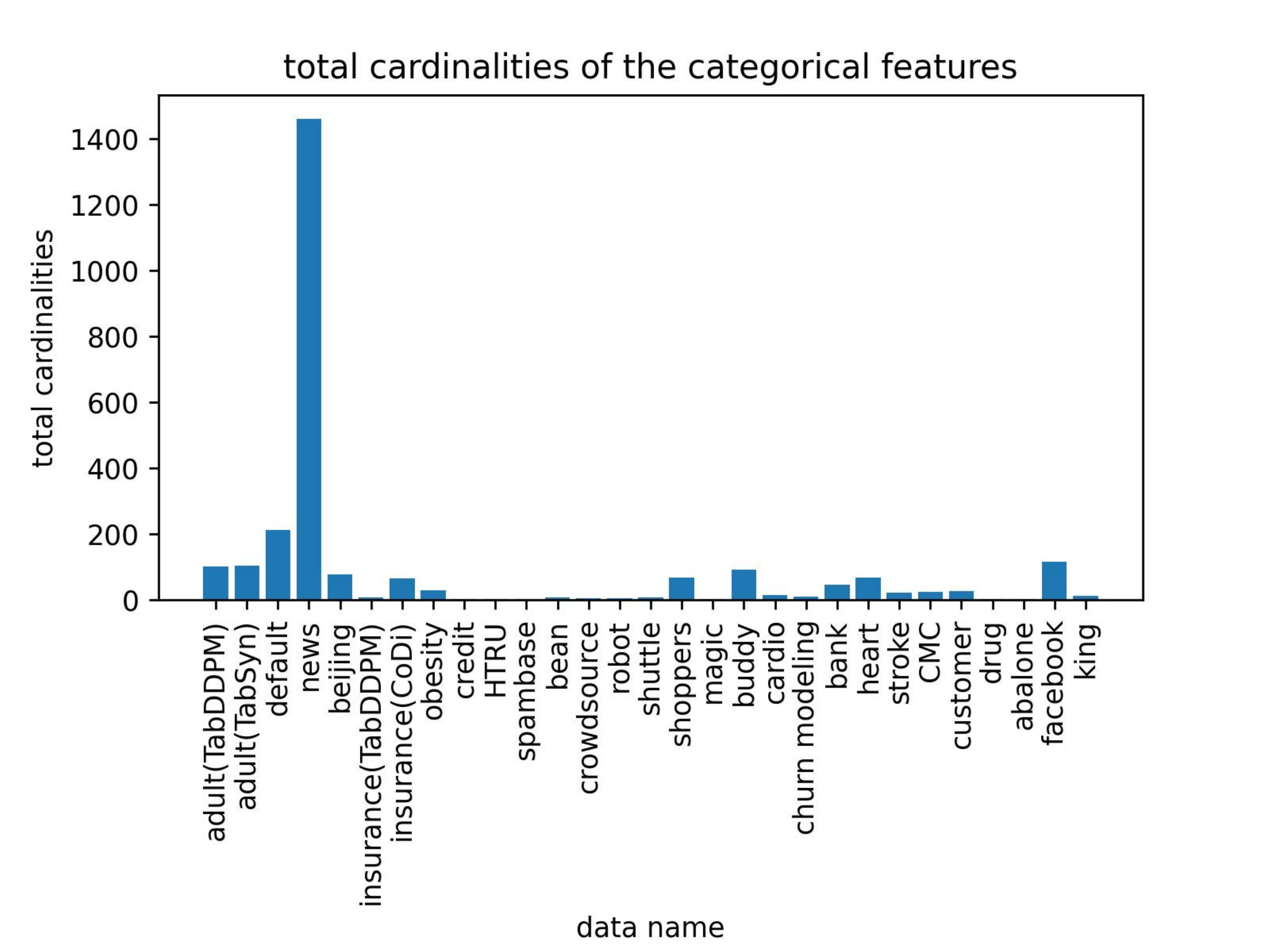}
  \caption{Plot of the total cardinalities of each data.}
  \label{fig:tot-cardinality}
  \vskip 0.1in
\end{figure}

In this paper, we focus on tabular data generation and consider the solution to the ``curse of dimensionality'' associated with one-hot vectors. The research on tabular data generation evaluates its performance using various datasets. Each paper provides information about the datasets used, including not only the data size but also details such as the type of task and the number of columns. However, in many cases, despite the conversion of categorical data to one-hot vectors, there has been a lack of examination regarding the increase in dimensionality. 
Figure~\ref{fig:tot-cardinality} presents our investigation into the cardinality of the categorical features for datasets used in recent tabular data generation research. This figure shows the total cardinalities for each data. The per-column cardinality of the each data is shown in Table~\ref{table:datasets-comparison}.

From this figure, it is evident that the cardinality is low for most datasets. When considering the application to real-world scenarios, we take the example of credit card transaction data. In such data, information about transactions includes details like `what was purchased?' and `where it was purchased?'. In typical scenarios with such datasets, the cardinality of categorical data is often extremely high~\footnote{For example, the zip code in the U.S. has 41,690 unique values.}. Considering the application of machine learning to real-world scenarios, verification in this aspect becomes essential.

To address the dimensionality increase of one-hot vectors accompanying the growth in cardinality, we propose \textbf{Res}idual \textbf{Bit} Vectors (ResBit) by integrating the concepts of Analog Bits~\cite{chen2023analog} and Residual Vector Quantization (RVQ)~\cite{1171604} (Section~\ref{subsec:resbit}). ResBit represents categorical data by hierarchically acquiring bit representations. Our research is motivated by the phenomenon of generation collapse observed in TabDDPM~\cite{tabddpm-pmlr-v202-kotelnikov23a}, one of the tabular data generative models. This phenomenon, akin to mode collapse, involves the generation of only specific values. This problem has also been encountered in other studies using TabDDPM as a baseline, as noted in the works of Zhang et al.~\yrcite{TabSyn-zhang2023mixedtype} and Shi et al.~\yrcite{shi2024tabdiffmultimodaldiffusionmodel}. While maintaining the same hyperparameter settings during the training of TabDDPM, we confirmed that the phenomenon of generation collapse is dependent on the cardinality of categorical data. This dependence is observed in conjunction with the dimensionality of the input to the multinomial diffusion models constituting TabDDPM, where the input is provided in the form of one-hot vectors. Based on this observation, we concluded that the dimensionality of one-hot vectors is a contributing factor to generation collapse. While conventional bit representations suffice for representing categorical data, they introduce specific problem in generation tasks (Section~\ref{subsec:out-of-index}). ResBit addresses this problem while mitigating the increase in dimensionality even with growing cardinality (Section~\ref{subsec:dimensions}).

To confirm the effectiveness of ResBit, we conduct experiments by applying it to TabDDPM (see Figure~\ref{fig:resbit}), which was the motivation for our research (Section~\ref{sec:exp}). Following the concept of analog bits, treating ResBit as numerical data allowed us to maintain performance in low cardinality datasets while generating diverse categorical values in high cardinality datasets. Additionally, we revealed advantages and disadvantages of existing tabular data generative models that have not been previously discussed. Furthermore, our experimental results contribute to providing answers to the question of how to handle categorical data in tabular data generation.

\section{Backgrounds}
\subsection{Tabular Data Generation with Deep Learning}
Generating tabular data using deep learning is highly meaningful. Unlike vision and NLP tasks, tabular data has limited data size and more directly includes personal information compared to text or images. Therefore, using synthetic data rather than raw data is an example of privacy consideration. TVAE and CTGAN~\cite{NEURIPS2019_254ed7d2} are models that apply VAE~\cite{kingma2013auto} and GAN~\cite{NIPS2014_5ca3e9b1} to tabular data generation, respectively, and are still used as baselines in the field. In addition to those, there is a distinct approach utilizing diffusion models~\cite{pmlr-v37-sohl-dickstein15, NEURIPS2020_4c5bcfec}. Research has been conducted on methods such as TabDDPM~\cite{tabddpm-pmlr-v202-kotelnikov23a}, CoDi~\cite{codi-pmlr-v202-lee23i}, STaSy~\cite{stasy-kim2023}, and TabSyn~\cite{TabSyn-zhang2023mixedtype}. These methods have demonstrated high performance compared to GAN-based approaches.

\subsection{Cardinality of the Categorical Features}
\label{subsec:cardinality-survey}
Figure~\ref{fig:tot-cardinality} shows the total cardinality of the each dataset used in recent tabular data generation researches. For more details, please refer to the Appendix~\ref{appendix:cardinality}. Many datasets shown in Figure~\ref{fig:tot-cardinality} have small cardinality for categorical features, making it challenging to assess the suitability of each method for real-world applications. For example, in e-commerce sites, an item ID may have high cardinality, and it is not demonstrated whether generation models can effectively handle such cases. IDs like these often represent entirely different entities with slight variations, making it difficult to treat them simply as continuous values. In this study, experiments are conducted on high cardinality datasets to assess the performance of existing methods.

\subsection{Discrete/Categorical Data in Diffusion Models}
\label{subsec:cat-diffusion}
The issue mentioned in Section~\ref{subsec:cardinality-survey} is similar to the treatment of tokens in text generation. Tokens, unlike semantic vectors, may represent entirely different meanings with slight variations in value. Consequently, research has been conducted on employing methods that handle discrete values, such as categorical data or text, with diffusion models originally designed for continuous values. D3PM~\cite{austin2021structured} and Multinomial Diffusion~\cite{hoogeboom2021argmax} are approaches that convert the noise distribution, initially assumed to be a normal distribution for continuous values, into a categorical distribution. Although the input is provided as a one-hot vector, these methods can learn imbalanced distributions as they predict categorical distributions. These techniques require designing models to adapt to categorical distributions. Bit Diffusion~\cite{chen2023analog} represents discrete data in binary and models analog bits, treated as continuous values, using Gaussian diffusion. It is characterized by the high expressiveness of diffusion models, eliminating the need for model designs tailored to discrete data. Diffusion-LM~\cite{li2022diffusionlm} is a method that learns diffusion models for word vectors, and similar ideas are applied to tabular data. TabSyn~\cite{TabSyn-zhang2023mixedtype} projects each column into a latent space and handles discrete quantities by processing them in continuous space. The treatment of categorical data involves two main strategies: treating it as a categorical distribution or converting it into continuous values. It is still unclear which approach is optimal. Our method treats categorical data as continuous values, representing a modification in the way these data are expressed. 

\section{Proposed Method}
\label{sec:proposed-method}
In this section, we first discuss the preliminaries and 
observation of the phenomenon of generation collapse in TabDDPM. Subsequently, we introduce the challenges of applying analog bits to tabular data generation tasks (Section~\ref{subsec:out-of-index}) and its solution, ResBit (Section~\ref{subsec:resbit}).

\subsection{Preliminaries}
\paragraph{TabDDPM}~\cite{tabddpm-pmlr-v202-kotelnikov23a} is the first study in the world to introduce diffusion models to tabular data generation. The tabular data consists of two types of data: numerical and categorical. In TabDDPM, numerical data is modeled using Gaussian diffusion models, while categorical data is modeled using multinomial diffusion~\cite{hoogeboom2021argmax}. The model is represented by MLP, and the coherence of the two types of data is maintained by integrating the losses. For numerical data, a simplified loss following DDPM~\cite{NEURIPS2020_4c5bcfec} is used as follows.

\begin{align*}
  L_t^{\mathrm{simple}}=\mathbb{E}_{t, \boldsymbol{x}_0, \varepsilon~\sim\mathcal{N}(0, \boldsymbol{I})}[\|\varepsilon-\varepsilon_{\theta}(\boldsymbol{x}_t, t)\|_2^2]
\end{align*}

Here, $\boldsymbol{x}_0$ denotes the real data and $\varepsilon_{\theta}$ denotes the neural network that predicts the noise $\varepsilon$ added at time step $t$. The categorical data also adopts the loss used in multinomial diffusion models. The predefined distributions for categorical data are represented as $q(\boldsymbol{x}_t|\boldsymbol{x}_{t-1})=\mathrm{Cat}(\boldsymbol{x}_t; (1-\beta_t)\boldsymbol{x}_{t-1}+\beta_t/K)$ when the number of classes is $K$. The loss functions for both types are combined to train the loss function of TabDDPM.

\begin{align*}
  L_{t}^{\mathrm{TabDDPM}}=L_t^{\mathrm{simple}}+\dfrac{\sum_{i\leq C}L_t^i}{C}
\end{align*}

$L_{t}^i$ represents two KL divergences between $q(\boldsymbol{x}_t| \boldsymbol{x}_{t-1})$ and $p_{\theta}(\boldsymbol{x}_{t-1}|\boldsymbol{x}_t)$ for each column.

\paragraph{Analog Bits}is a preprocessing method proposed for handling discrete data in Bit Diffusion~\cite{chen2023analog}. Normally, discrete data is represented as a one-hot vector, but analog bits represents it as bit strings. Furthermore, by treating bit strings as numerical data, continuous state diffusion models can be applied to discrete data without devising a specific architecture for discrete data.

\subsection{Generation Collapse in TabDDPM}
Table~\ref{table:collapse} shows whether a mode collapse-like phenomenon occurred when we randomly sampled 55,000 cases from IBM’s synthetic credit card data\footnote{\url{https://ibm.ent.box.com/v/tabformer-data}} and changed the number of unique values for the data of each category. The percentage here is the percentage of the total training data, and data with frequencies lower than this percentage are masked\footnote{This is represented as \texttt{cat\_min\_frequency} in the official TabDDPM implementation.}. Although the dataset used here is only a partial selection, it contains a very large number of categorical values. In this experiment, the hyperparameters except the value indicating the masking ratio are fixed, so changes in this ratio are linked to changes in the dimensionality of the one-hot vector given as input to the multinomial diffusion models. On the basis of this experiment, we hypothesize that the increase in the dimensionality of the one-hot vector affects the success or failure of training.

\begin{table}[ht]
\caption{Wether collapse does occur or not when the percentage of masking was changed in the experiment}
\label{table:collapse}
\begin{center}
\begin{small}
\begin{tabular}{cc}
\toprule
Threshold (Rate) &  collapsed?  \\
\midrule
0.00038 & No \\
0.00028 & Yes \\
0.00019 & Yes \\
0.00010 & Yes \\
\bottomrule
\end{tabular}
\end{small}
\end{center}
\end{table}

For more details, we discuss it using simple synthetic data in Appendix~\ref{appendix:toy-data}.

In terms of the formulation of multinomial diffusion, the forward process of multinomial diffusion is formulated using a categorical distribution as follows:

\begin{align}
  & q(\boldsymbol{x}_t|\boldsymbol{x}_{t-1})\coloneqq\mathrm{Cat}(\boldsymbol{x}_t; (1-\beta_t)\boldsymbol{x}_{t-1}+\beta_t/K\cdot\boldsymbol{1}) \\
  & q(\boldsymbol{x}_T)\coloneqq\mathrm{Cat}(\boldsymbol{x}_T; 1/K\cdot\boldsymbol{1}) \\
  & q(\boldsymbol{x}_t|\boldsymbol{x}_0)=\mathrm{Cat}(\boldsymbol{x}_t|\overline{\alpha}_t\boldsymbol{x}_0+(1-\overline{\alpha}_t)/K\cdot\boldsymbol{1})
\end{align}

where $\alpha_t\coloneqq 1-\beta_t, \overline{\alpha}_t=\prod_{i\leq t}\alpha_i$. $\mathrm{Cat}$ denotes the categorical distribution and $K$ denotes the number of classes. $\beta_t$ determines the strength of added noise and satisfies $0<\beta_1<\dotsi<\beta_T<1$. From the equations above, the posterior $q(\boldsymbol{x}_{t-1}|\boldsymbol{x}_t, \boldsymbol{x}_0)$ can be expressed:

\begin{align}
q(\boldsymbol{x}_{t-1}|\boldsymbol{x}_t, \boldsymbol{x}_0)=\mathrm{Cat}\left(\boldsymbol{x}_{t-1}\mid \boldsymbol{\pi}/\sum_{j=1}^K\pi_j\right)
\end{align}

where $\boldsymbol{\pi}=[\alpha_t\boldsymbol{x}_t+(1-\alpha_t)/K\cdot\boldsymbol{1}]\odot[\overline{\alpha}_{t-1}\boldsymbol{x}_0+(1-\overline{\alpha}_{t-1})/K\cdot\boldsymbol{1}]$.

We consider a high cardinality scenario, that is, $K\to\infty$. Since $\boldsymbol{x}_0$ is represented as a one-hot vector, we have $x_{0, k}=1$ and $x_{0, j}=0$ for $j\neq k$.

\begin{align}
\lim_{K\to\infty}\boldsymbol{\pi}&=[\alpha_t\boldsymbol{x}_t]\odot[\overline{\alpha}_{t-1}\boldsymbol{x}_0] \nonumber \\
&=(\alpha_t x_{t, i}\cdot \overline{\alpha}_{t-1}x_{0, i})_{0\leq i\leq K}=\alpha_{t}x_{t, k}\cdot\overline{\alpha}_{t-1}  
\end{align}

Therefore,

\begin{align}
\lim_{K\to\infty}\boldsymbol{\pi}/\sum_{j=1}^K\pi_j=\boldsymbol{x}_0
\end{align}

Then,

\begin{align}
q(\boldsymbol{x}_{t-1}|\boldsymbol{x}_t, \boldsymbol{x}_0)=\mathrm{Cat}(\boldsymbol{x}_{t-1}|\boldsymbol{x}_0)
\end{align}

and we are considering a situation where only a specific class appears.

\subsection{Out of Index}
\label{subsec:out-of-index}
\begin{figure}[htbp]
  \vskip 0.1in
  \centering
  \includegraphics[width=0.8\linewidth]{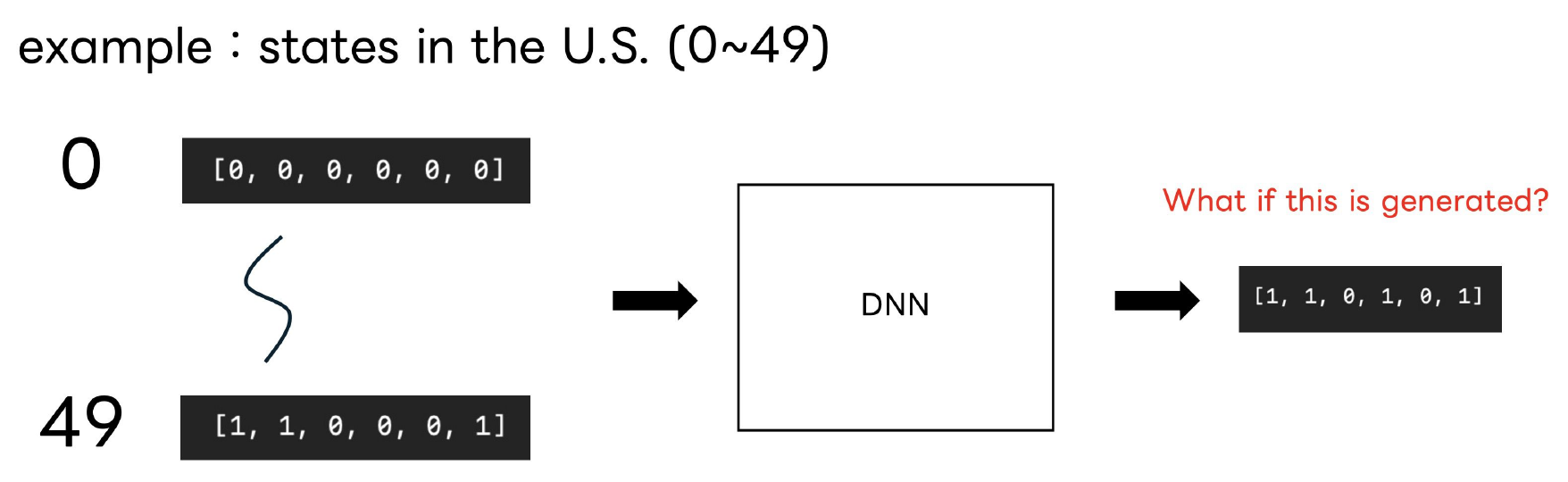}
  \caption{Overview of the ``out of index'' problem in the states of the U.S. example.}
  \label{fig:ooi}
  \vskip 0.1in
\end{figure}

In binary representation, there is a possibility of generating values that do not exist in the generation task. As an example of this problem, let's consider representing the states in the United States. There are 50 states in the United States, and if we represent them using one-hot vectors, it would be like $\boldsymbol{x}\in\{0, 1\}^{50}$. Specifically, for $j$-th state, $x_j=1$ and $x_i=0$ for $i\neq j$. Representing this in binary would be possible with a 6-dimensional representation. The model generates each element as either 0 or 1, resulting in $2^6=64$ possible combinations. However, out of these combinations, 14 would represent state numbers that do not correspond to any existing states. For example, if the model predicts $(1, 1, 0, 1, 0, 1)$, it corresponds to the number 53, but this value cannot be mapped to any existing state in the context of representing the 50 states of the United States. We conveniently refer to this problem as \textbf{out of index}, and an overview is provided in Figure~\ref{fig:ooi}. Increasing the number of trainings could potentially address this problem. However, this approach is effective when the training data is extensive, and for tabular data generation, which involves training on individual datasets, there is a simple risk of overfitting.

\subsection{Residual Bit Vectors}
\label{subsec:resbit}
A straightforward solution to address the ``out of index'' problem discussed in Section~\ref{subsec:out-of-index} is to limit categorical data to one special string and $2^n-1$ different categories. However, this method imposes limitations on the inherent diversity of the training data. Other approaches such as clipping or normalizing out-of-range values are also conceivable. However, the clipping method introduces issues where the distribution of generated data may change. In the case of normalization, as mentioned in Section~\ref{subsec:cat-diffusion}, there is a risk that slight value differences might refer to entirely different entities, making this solution highly inappropriate. In essence, the maximum representable number should be the number of categorical data. In the example of U.S. states mentioned earlier, as there are 50 states, a representation method that indicates integers from 0 to 49 inclusive would be necessary.

\begin{figure}[htbp]
  \vskip 0.1in
  \centering
  \includegraphics[width=0.7\linewidth]{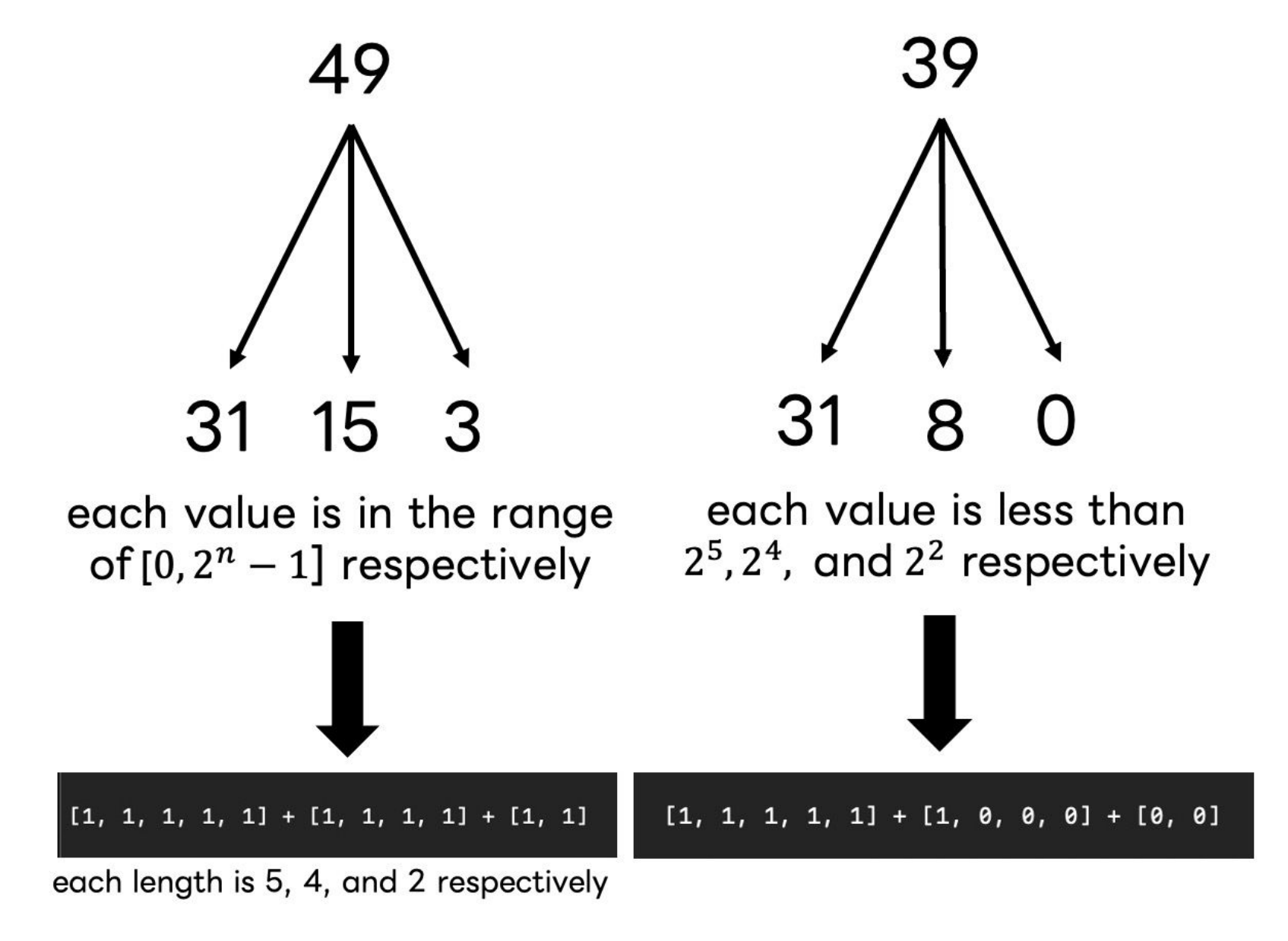}
  \caption{ResBit in the states of the U.S. example.}
  \label{fig:resbit-us}
  \vskip 0.1in
\end{figure}

We propose \textbf{Res}idual \textbf{Bit} Vectors (ResBit) inspired by RVQ to fulfill the aforementioned requirement. RVQ recursively applies vector quantization to the quantization error obtained when quantizing the original vector. ResBit accomplishes this using a binary representation. Figure~\ref{fig:resbit-us} provides an overview of ResBit. When using ResBit, like one-hot vectors, prior knowledge of the number of classes is required. Let $M$ represent the number of classes. There exists a non-negative integer $b_1$ such that inequality (\ref{eq:m-ineq}) is satisfied. Representing $2^{b_1}-1$ in binary results in a bit sequence where all elements are 1s, and its length is $b_1$.

\begin{equation}
\label{eq:m-ineq}  
2^{b_1}-1\leq M-1<2^{b_1+1}-1
\end{equation}

In this state, it can represent values from 0 to $2^{b_1}-1$, but cannot represent values outside this range. Therefore, the same process is applied to $M-(2^{b_1}-1)$. This repetition is how the representation of ResBit is obtained. Formally, the required ResBit for representing $M$ classes, indexed from 0, can be written as follows:

\begin{equation}
\label{eq:resbit-dim}
\min\sum_{k=0}^\infty a_kk \quad \mathrm{s.t.}\quad M-1=\sum_{k=0}^\infty a_k(2^k-1)
\end{equation}

The $k$-th term represents integers in the range $[0, a_k(2^k-1)]$. An integer $N\in\mathbb{Z}$, where $0\leq N<M$, can be represented as follows:

\begin{equation}
  N=\sum_{k=1}^\infty c_k\quad (c_k\in[0, a_k(2^k-1)], c_k\in\mathbb{Z})
\end{equation}

Finally, we present the example of U.S. states mentioned earlier. Since $M=50$, we have $50-1=49=1\cdot(2^5-1)+1\cdot(2^4-1)+1\cdot(2^2-1)=31+15+3$. In other words, U.S. states can be represented in $11(=5+4+2)$ dimensions. Using this approach, we can represent all integers from 0 to 49 inclusive. For example, for the case of 39, which is $39=31+8$, the ResBit representation is $39=((1,1,1,1,1),(1,0,0,0),(0,0))=(1,1,1,1,1,1,0,0,0,0,0)$. This example is illustrated in Figure~\ref{fig:resbit-us}. Pseudo-code for ResBit is provided in Appendix~\ref{appendix:resbit-impl}.

\subsection{Decreasing the Dimensionality}
\label{subsec:dimensions}
\begin{figure}[htbp]
  \vskip 0.1in
  \centering
  \includegraphics[width=0.7\linewidth]{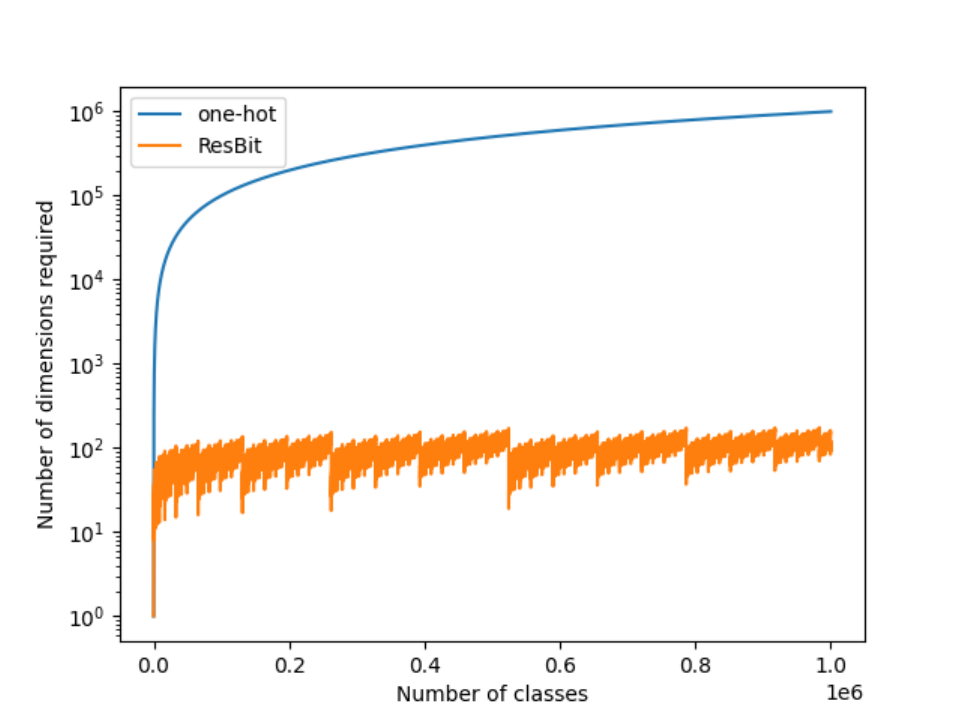}
  \caption{Comparison of the dimensionality between one-hot and ResBit when the number of classes are increased to $10^6$.}
  \label{fig:comp-dim}
  \vskip 0.1in
\end{figure}

Due to its properties, ResBit can reduce the increase in dimensionality compared to one-hot vectors. Figure~\ref{fig:comp-dim} illustrates the comparison of the required dimensionality for representing integers less than $M$ when the number of classes is $M$. For example, when $M=1,000$, the required dimensionality is 1,000 for one-hot vectors, while ResBit only requires 42 dimensions, resulting in a reduction of 95.8\%. This allows us to alleviate the ``curse of dimensionality'' associated with using one-hot vectors.

\section{Experiments}
\label{sec:exp}
We consider Table Residual Bit Diffusion (TRBD), which is incorporated ResBit into TabDDPM (see Figure~\ref{fig:resbit}). In this section, we demonstrate the effectiveness of integrating ResBit into TabDDPM through experiments. When introducing ResBit into TabDDPM, it is challenging to directly apply ResBit as multinomial diffusion assumes one-hot vectors as inputs. Therefore, following the concept of analog bits, ResBit is treated as numerical data. ResBit is combined with numerical data, and preprocessing is performed using \texttt{QuantileTransformer}. Specifically, One-Hot Encoder is replaced by Residual Bit Encoder, and its outputs are treated as numerical data following the idea of analog bits. After converting each categorical value to its ResBit, the process described in~\cref{eq:preprocess} is performed. This is the same process as that of TabDDPM.

\begin{equation}
\label{eq:preprocess}
  x_{cat_i}^{resbit}=\log(\max(\texttt{ResBit\_Encoder}(x_{cat_i}), 10^{-30}))
\end{equation}

Here, \texttt{ResBit\_Encoder} converts the category value to ResBit, and $x_{cat_i}$ is the $i$-th categorical column. After converting categorical values to ResBit, we combine them with numerical data and further transform them using Quantile Transformer.

\begin{equation*}
  x_{in}=\mathrm{QuantileTransformer}(\mathrm{concat}(x_{num}, x_{cat_1}^{resbit}, \ldots))
\end{equation*}

\subsection{Experimental Setups}
We briefly describe the baselines, datasets, and evaluation methods. For software information and further details, please refer to Appendices~\ref{appendix:exp-setting},~\ref{appendix:baselines}, and~\ref{appendix:datasets}.
\paragraph{Baselines.} 
We use six existing methods as baselines. The first two are CTGAN~\cite{NEURIPS2019_254ed7d2} and TVAE~\cite{NEURIPS2019_254ed7d2} as the de fact standards. Additionally, we include TabDDPM~\cite{tabddpm-pmlr-v202-kotelnikov23a}, CoDi ~\cite{codi-pmlr-v202-lee23i}, StaSy~\cite{stasy-kim2023}, and TabSyn~\cite{TabSyn-zhang2023mixedtype} as the diffusion-based methods. These methods were nearly simultaneously introduced, and comprehensive comparisons have not been extensively conducted. Zhang et al.~\yrcite{TabSyn-zhang2023mixedtype} conducted a comparison, but the evaluation was based on only six datasets, which is limited compared to 11 datasets~\cite{codi-pmlr-v202-lee23i} or 15 datasets~\cite{tabddpm-pmlr-v202-kotelnikov23a, stasy-kim2023}.
\paragraph{Datasets.}
We use 10 datasets to conduct a comprehensive evaluation of the existing methods and verify the effectiveness of our proposed method. The basic overview of the dataset is presented in Table~\ref{table:simple-datasets}. For additional details such as the number of training data, please refer to Table~\ref{table:datasets} in Appendix~\ref{appendix:datasets}. 
ResBit is a method representing categorical data. Therefore, experiments are conducted on datasets that include both categorical and numerical data. To validate the effectiveness of our proposed method, we prepare datasets with both low and high cardinalities. Low cardinality datasets primarily involve datasets used in TabDDPM~\cite{tabddpm-pmlr-v202-kotelnikov23a}. There are \textbf{three main purposes} for evaluating on datasets of these two different scales. 
\begin{enumerate}
  \item Comparing the performance of the existing methods as a comprehensive evaluation
  \item Verifying the performance of our proposed method on low cardinality datasets (maintaining the performance of TabDDPM)
  \item Assessing the effectiveness of our proposed method and confirming the performance of the existing methods on high cardinality datasets
\end{enumerate}
Most datasets are real-world datasets, except for the CC dataset, which is a synthetic dataset.

\begin{table}[!ht]
\caption{Brief overview of the datasets used in our experiments.}
\label{table:simple-datasets}
\vskip 0.1in
\begin{center}
\begin{footnotesize}
\begin{tabular}{lcc}
\toprule
Dataset Name & Abbreviation & Cardinality \\
\midrule
Credit Card (IBM) & CC & \multirow{2}{*}{high} \\
Airlines & AR & \\
\midrule
Insurance & IS & \multirow{8}{*}{low} \\
Buddy & BD &  \\
Adult & AD &  \\
Churn Modeling & CH & \\
Cardio & CA & \\
King & KI & \\
Abalone & AB & \\
Facebook Comm. Vol. & FB & \\
\bottomrule
\end{tabular}
\end{footnotesize}
\end{center}
\end{table}

\paragraph{Evaluation methods}
We evaluate the existing and our methods from three aspects. 1) \textbf{cardinality of the categorical features}: compare the number of generated categories in each categorical column, aiming for a metric that reflects proximity to the training data. 2) \textbf{TSTR} (Train in Synthesis, Test on Real)~\cite{esteban2017realvalued, yoon2018pategan}: train a classifier on generated data and evaluate its performance on test data to measure sampling quality. We use CatBoost~\cite{NEURIPS2018_14491b75} as the classifier, and conduct five measurements, reporting the average and standard deviation. Experiments are conducted with 10,000 samples for the low cardinality datasets and 3,000,000 samples for the high cardinality datasets. 3) \textbf{runtime}: confirm the impact of applying ResBit on the runtime.

\subsection{TSTR Framework}
\label{subsec:tstr-exp}
We show the results in Tables \ref{table:tstr-result-auroc} and \ref{table:tstr-result-r2}. Additional results can be found in Appendix~\ref{appendix:additional-results}. \texttt{Failed} indicates that the evaluation using TSTR was not possible due to insufficient diversity in the generated labels for the classification task. In addition, if the training did not complete within 24 hours, the result is marked as \texttt{N/A}, adhering to the criteria outlined in the generative trilemma~\cite{xiao2022tackling}. The generative trilemma outlines speed, quality, and diversity as the three requirements for the sampling stage in generative tasks. However, we argue that these criteria should also be applicable during the training phase. The ideal scenario for a generative task is to achieve diverse and high-quality generation rapidly through short training. Therefore, we believe that investing time in training solely to meet the requirements of the generative trilemma is inappropriate.

\begin{table*}[!htbp]
\caption{Results of TSTR framework in classification datasets. Metric is AUROC.}
\label{table:tstr-result-auroc}
\vskip 0.15in
\begin{center}
\begin{small}
\begin{tabular}{lcc|cccc}
\toprule
& CC & AR  & BD & AD & CH & CA \\
\midrule
Real     
& .971{\tiny$\pm$.009} % CC
& .727{\tiny$\pm$.004} % AR
& .931{\tiny$\pm$.001} % BD
& .927{\tiny$\pm$.002} % AD
& .855{\tiny$\pm$.010} % CH
& .805{\tiny$\pm$.000} % CA
\\ \midrule
TVAE
& \texttt{Failed}
& .640{\tiny$\pm$.002}
& .978{\tiny$\pm$.001}
& .865{\tiny$\pm$.004}
& .763{\tiny$\pm$.012}
& .641{\tiny$\pm$.010} \\
CTGAN
& \texttt{Failed}
& .639{\tiny$\pm$.009}
& .955{\tiny$\pm$.004}
& .874{\tiny$\pm$.002}
& .674{\tiny$\pm$.015}
& .749{\tiny$\pm$.007} \\
TabDDPM  
& .376{\tiny$\pm$.075}
& .528{\tiny$\pm$.013}
& \textbf{.984{\tiny$\pm$.001}}
& \textbf{.909{\tiny$\pm$.003}}
& \textbf{.873{\tiny$\pm$.002}}
& \textbf{.803{\tiny$\pm$.001}} \\
CoDi     
& \texttt{N/A} 
& \texttt{N/A} 
& .472{\tiny$\pm$.039} 
& .495{\tiny$\pm$.062} 
& .584{\tiny$\pm$.017} 
& .497{\tiny$\pm$.034} \\
STaSy    
& \texttt{N/A} 
& \texttt{N/A} 
& .978{\tiny$\pm$.002} 
& .900{\tiny$\pm$.002} 
& .835{\tiny$\pm$.004} 
& .791{\tiny$\pm$.001} \\
TabSyn   
& \textbf{.930{\tiny$\pm$.012}}
& \textbf{.690{\tiny$\pm$.001}}
& \textbf{.984{\tiny$\pm$.001}}
& .899{\tiny$\pm$.003}
& .839{\tiny$\pm$.003}
& .791{\tiny$\pm$.001} \\
\midrule
TRBD (ours) 
& .746{\tiny$\pm$.051}
& .638{\tiny$\pm$.003}
& \textbf{.983{\tiny$\pm$.001}}
& .906{\tiny$\pm$.000}
& \textbf{.870{\tiny$\pm$.002}}
& \textbf{.803{\tiny$\pm$.001}} \\
\bottomrule
\end{tabular}
\end{small}
\end{center}
\vskip 0.1in
\end{table*}

\begin{table*}[!htbp]
\caption{Results of TSTR framework in regression datasets. Metric is $R^2$ score.}
\label{table:tstr-result-r2}
\vskip 0.15in
\begin{center}
\begin{small}
\begin{tabular}{lccccc}
\toprule
& IS & KI & AB & FB \\
\midrule
Real     
& .861{\tiny$\pm$.034} % IS
& .906{\tiny$\pm$.002} % KI
& .557{\tiny$\pm$.002} % AB
& .838{\tiny$\pm$.001} % FB
\\ \midrule
TVAE
& .634{\tiny$\pm$.026}
& .650{\tiny$\pm$.002}
& .278{\tiny$\pm$.006}
& .539{\tiny$\pm$.009} \\
CTGAN
& $-0.706${\tiny$\pm$.095}
& .628{\tiny$\pm$.012}
& .291{\tiny$\pm$.029}
& .420{\tiny$\pm$.012} \\
TabDDPM  
& .894{\tiny$\pm$.003}
& .864{\tiny$\pm$.112}
& \textbf{.567{\tiny$\pm$.012}}
& .662{\tiny$\pm$.008} \\
CoDi     
& $-4.506${\tiny$\pm$.616} 
& $-88.074${\tiny$\pm$20.908} 
& $-2.518${\tiny$\pm$.414} 
& \texttt{N/A} \\
STaSy    
& .618{\tiny$\pm$.080} 
& .867{\tiny$\pm$.010} 
& $-0.613${\tiny$\pm$.047} 
& .670{\tiny$\pm$.011} \\
TabSyn   
& .834{\tiny$\pm$.004} 
& .867{\tiny$\pm$.005} 
& .522{\tiny$\pm$.005} 
& .616{\tiny$\pm$.009} \\
\midrule
TRBD (ours) 
& \textbf{.908{\tiny$\pm$.003}}
& \textbf{.892{\tiny$\pm$.004}}
& .560{\tiny$\pm$.006}
& \textbf{.706{\tiny$\pm$.005}} \\
\bottomrule
\end{tabular}
\end{small}
\end{center}
\vskip 0.1in
\end{table*}

\paragraph{Comparison of the baselines.}
We observe that CoDi requires a substantial amount of time for training when the size of the training data increases. The size of the datasets that Lee et al.~\yrcite{codi-pmlr-v202-lee23i} used is up to 36k. However, the size of the FB dataset is 157k. Therefore, it can be inferred that the impact of the dataset size on training time could not be identified. As evident from Table~\ref{table:datasets-comparison}, the cardinality of the datasets used by Kim et al.~\yrcite{stasy-kim2023} is low, making it difficult to assess the impact of cardinality on training time. However, the experiments reveal that high cardinality datasets indeed require a significant amount of time for training. Indeed, as demonstrated, existing methods may not perform well on high cardinality datasets, an aspect that has not been widely recognized until now. 

Roughly, similar results to Zhang et al.~\yrcite{TabSyn-zhang2023mixedtype} are obtained, but in our experiments, TabDDPM consistently shows superior performance, suggesting the need for further extensive validation. Additionally, in the large-scale datasets (CC and AR), TabSyn demonstrates superiority. This might be attributed to the Transformer component in TabSyn, which is effective with large scale datasets, possibly related to the requirements of Vision Transformer~\cite{dosovitskiy2021an}. Transformers are known to adhere to Scaling Laws~\cite{kaplan2020scaling}, making them more effective with larger datasets. On the other hand, CTGAN outperforms in F1 Score on the AR dataset according to Tables~\ref{table:cat-f1}, \ref{table:xgb-f1}, and \ref{table:rf-f1}. However, CTGAN fails to generate effectively in the highly imbalanced CC dataset. Thus, it is challenging to conclusively state the superiority of GAN-based methods based on these results.
\paragraph{Evaluation of our proposed method.}
From Tables~\ref{table:tstr-result-auroc} and \ref{table:tstr-result-r2}, it can be observed that TRBD maintains performance across most datasets compared to TabDDPM. Moreover, considering the results in Tables~\ref{table:cat-f1}-\ref{table:rf-rmse}, an overall improvement in performance is evident. In particular, significant performance improvement is observed in the high cardinality datasets (CC and AR). This indicates that multinomial diffusion struggles to handle high cardinality datasets effectively. Furthermore, the reduced standard deviations suggest stable and consistent sampling quality. 
\paragraph{Overall discussion.}
Taking into account Tables~\ref{table:cat-f1}-\ref{table:beta-recall} in addition to Tables \ref{table:tstr-result-auroc} and \ref{table:tstr-result-r2}, diffusion models, specifically methods treating categorical data as continuous (TRBD, STaSy, and TabSyn), show a tendency for high sampling quality. In particular, methods that handle categorical and numerical data separately, such as TabDDPM, have not shown competitive results on high cardinality datasets. Additionally, CoDi performs poorly overall. In contrast, STaSy, TabSyn, and TRBD, which process categorical data within the same framework as numerical data, demonstrate competitive results on many datasets. TRBD and TabSyn, which employ dimensionality reduction techniques, also excel on high cardinality datasets, showcasing their effectiveness.

\subsection{Cardinality of the Categorical Features}
\label{subsec:exp-cardinality}
Table~\ref{table:cardinarity-rate} shows a comparison of the cardinality of the categorical features. We confirm this by generating an equal number of instances with each generative model. Due to datasets with a high number of columns, it is not possible to show the results for each column. Therefore, we present the ratio to the cardinality of the training data, indicating the maximum ratio from five generations. In low cardinality datasets, the results are similar, whereas in high cardinality datasets, the differences are pronounced. If the models can generate diverse samples, this ratio should approach 1. CoDi seems to produce diverse results, but from Tables~\ref{table:tstr-result-auroc} and~\ref{table:tstr-result-r2}, it can be argued that the sampling quality is not good. Both CTGAN and TVAE show a loss of diversity. Neither of them can be considered to achieve the generative trilemma. The lack of diversity is also observed in TabSyn. This suggests the potential loss of fine-grained information due to handling data in the latent space.

\begin{table*}[!htbp]
\caption{Comparison of the cardinality of the categorical features in the generated samples.}
\label{table:cardinarity-rate}
\vskip 0.15in
\begin{center}
\begin{small}
\begin{tabular}{lcc|cccccccc|c}
\toprule
& CC & AR & IS & BD & AD & CH & CA & KI & AB & FB & Average \\
\midrule
TVAE    & 0.526 & 0.917 & 1.000 & 0.576 & 0.618 & 0.778 & 0.929 & 1.000 & 1.000 & 0.862 & 0.821 \\
CTGAN   & 0.369 & 0.764 & 1.000 & 1.000 & 1.000 & 1.000 & 1.000 & 1.000 & 1.000 & 0.899 & 0.903 \\
TabDDPM & 0.487 & 0.832 & 1.000 & 0.870 & 1.000 & 1.000 & 1.000 & 1.000 & 1.000 & 0.973 & 0.916 \\
CoDi    & \texttt{N/A} & \texttt{N/A} & 1.000 & 1.000 & 1.000 & 1.000 & 1.000 & 1.000 & 1.000 & \texttt{N/A} & 1.000+3\texttt{N/A} \\
STaSy   & \texttt{N/A} & \texttt{N/A} & 1.000 & 0.924 & 0.990 & 1.000 & 1.000 & 1.000 & 1.000 & 0.963 & 0.985+2\texttt{N/A} \\
TabSyn  & 0.910 & 1.000 & 1.000 & 0.815 & 0.980 & 1.000 & 1.000 & 1.000 & 1.000 & 1.000 & 0.971 \\
\midrule
TRBD (ours) & 1.000 & 1.000 & 1.000 & 0.957 & 1.000 & 1.000 & 1.000 & 1.000 & 1.000 & 1.000 & \textbf{0.996} \\
\bottomrule
\end{tabular}
\end{small}
\end{center}
\vskip 0.1in
\end{table*}

\subsection{Runtime}
\label{subsec:exp-runtime}
Table~\ref{table:runtime} shows the comparison of the runtime. We use the CH dataset and the CC dataset. The experimental setup has the training batch size of 4,096 and the number of generated samples is 26k for the CH dataset and 100k for the CC dataset. Sampling batch size is equal to the number of generated samples. Other values have not been changed from previous experiments in order to maintain the sampling quality. TabSyn’s training time is the summation of VAE’s and denoising network’s training time. Moreover, note that TabSyn’s denoising network is equipped with early stopping, so training ends early.

\begin{table*}[htbp]
\caption{Comparison of the runtime. \texttt{OOM} denotes out of memory with 24GB GPU.}
\label{table:runtime}
\vskip 0.15in
\begin{center}
\begin{small}
\begin{tabular}{lccccc}
\toprule
& \multicolumn{2}{c}{CH} & \multicolumn{2}{c}{CC} \\
Method  & Training & Sampling & Training & Sampling \\
\midrule
TVAE    & \textbf{33.01}s & \textbf{0.07}s & 1,973s & 33.22s \\
CTGAN   & 33.06s & 0.22s & 36,797s & 225s \\
TabDDPM & 567s & 22.78s & 3,494s & \texttt{OOM} \\
CoDi    & 6,149s & 1.61s & \texttt{N/A} & \texttt{N/A} \\
STaSy   & 4,197s & 10.40s & \texttt{N/A} & \texttt{N/A} \\
TabSyn  & 2,325s$+$488s & 4.34s & 27,377s+2,737s & 15.09s \\
\midrule
TRBD (ours) & 264s & 13.57s & \textbf{183}s & \textbf{4.31}s \\
\bottomrule
\end{tabular}
\end{small}
\end{center}
\vskip 0.1in
\end{table*}

In the CH dataset, TRBD achieves approximately twice the speedup compared to TabDDPM. Moreover, for the CC dataset, it accomplishes a speedup of almost 20 times during the training phase. TabDDPM and TRBD have the same number of training iterations. Therefore, it is suggested that the computational cost of the multinomial diffusion part in TabDDPM becomes a bottleneck. In other words, considering runtime, TRBD is faster than TabDDPM, capable of generating a wide range of data, and achieving comparable or superior quality. STaSy takes considerably more time, especially for low cardinality datasets, when compared to TabDDPM, TabSyn, and TRBD. This becomes even more pronounced for high cardinality datasets. While the sampling is performed rapidly, satisfying the generative trilemma, the training process is time-consuming. Although the generated quality is competitive, the training time serves as a bottleneck in achieving it. TabDDPM using one-hot vectors demands a significant amount of memory, particularly on the CC dataset. In contrast, TRBD does not exhibit such memory-intensive behavior, resulting in reduced computational costs.

\section{Conclusion and Future Work}
In this paper, we focus on the mode collapse phenomenon in TabDDPM, investigate its cause, and propose ResBit, a hierarchical bit representation, as a solution to avoid it. ResBit is an extension of analog bits to address the ``out-of-index'' problem. This hierarchical bit representation theoretically avoids the ``out-of-index'' problem. By treating ResBit as numerical data when applying it to tabular data generative models, training time is shortened because of the reduction in the dimensionality of categorial data, the sampling quality is maintained, and diverse categorical values are generated. Throughout our comprehensive experiments, we also observe that a multitude of existing methods do not demonstrate the effective results in the settings not explored in previous research such as high cardinality dataset and the large-scale dataset.

Although we focus on cardinalities, we believe there are other important factors, which have not received sufficient attention in tabular data generation. In addition, since the results are dependent on the dataset, further validation with a wider range of data is also required.

% In the future, further validation of ResBit is desired. For instance, although we confirmed its effectiveness in image classification and class-conditional image generation tasks in Appendix~\ref{appendix:others}, the experiments were not exhaustive, and there are other potential applications, such as image segmentation or algorithms utilizing one-hot vectors. Furthermore, as existing methods for tabular data generation all convert categorical data into one-hot vectors, constructing models with fewer parameters becomes possible by applying ResBit. While we demonstrated its application to TabDDPM in this paper, its effectiveness is likely to extend to other models as well.

\bibliography{example_paper}
\bibliographystyle{icml2024}

%%%%%%%%%%%%%%%%%%%%%%%%%%%%%%%%%%%%%%%%%%%%%%%%%%%%%%%%%%%%%%%%%%%%%%%%%%%%%%%
% APPENDIX
%%%%%%%%%%%%%%%%%%%%%%%%%%%%%%%%%%%%%%%%%%%%%%%%%%%%%%%%%%%%%%%%%%%%%%%%%%%%%%%
\newpage
\appendix
\onecolumn
\section{Details of Experimental Setups}
\subsection{Experimental Environments}
\label{appendix:exp-setting}
Our software and hardware environments are as follows: Ubuntu 18.04, Python 3.10.8, PyTorch 1.13.1, scikit-learn 1.1.2, and CUDA 10.1, AMD EPYC 7502P 32-Core CPU at 2.5GHz, and one Quadro RTX 6000.

\subsection{Baselines}
\label{appendix:baselines}
We describe the baseline methods used in our experiments. The baseline methods consist of various types of generative models, including gan-based methods, vae-based methods, and diffusion-based methods.

\begin{itemize}
  \item \textbf{CTGAN} and \textbf{TVAE}~\cite{NEURIPS2019_254ed7d2}\footnote{\url{https://github.com/sdv-dev/CTGAN}}: The same technique has been applied in various studies to both GANs and VAEs, where it serves as a common baseline in research.
  \item \textbf{TabDDPM}~\cite{tabddpm-pmlr-v202-kotelnikov23a}\footnote{\url{https://github.com/yandex-research/tab-ddpm}}: This study represents the first introduction of diffusion models for tabular data generation, employing separate models for numerical and categorical data. A shared MLP network is utilized, and learning is achieved by separating the losses. Gaussian diffusion models are employed for numerical data, while multinomial diffusion models are used for categorical data.
  \item \textbf{CoDi}~\cite{codi-pmlr-v202-lee23i}\footnote{\url{https://github.com/ChaejeongLee/CoDi/tree/main}}: Similar to TabDDPM, numerical data is learned using Gaussian diffusion, and categorical data is learned using multinomial diffusion. However, in this case, two separate networks are employed for learning. The relationships between columns are learned by conditioning the output of each model on the other. Additionally, the use of symmetric learning methods is introduced to strengthen connections. 
  \item \textbf{STaSy}~\cite{stasy-kim2023}\footnote{\url{ https://github.com/JayoungKim408/STaSy/tree/main}}: This diffusion-based generative model employs variance exploding, variance preserving, and sub-variance preserving SDEs~\cite{song2021scorebased} to train the distribution of the tabular data. By incorporating self-placed learning, a curriculum learning technique, and fine-tuning, the method stabilizes the quality of generated samples.
  \item \textbf{TabSyn}~\cite{TabSyn-zhang2023mixedtype}\footnote{\url{https://github.com/amazon-science/TabSyn}}: This is inspired by Latent Diffusion Models~\cite{Rombach_2022_CVPR}. It employs a transformer architecture VAE for projecting into the latent space and an MLP for denoising. The training is designed to embed both numerical and categorical data into the latent space while maintaining their associations, enabling the generation of horizontal connections specific to tabular data. Furthermore, performing denoising in the latent space makes it faster compared to existing diffusion-based methods.
\end{itemize}

We use official implementations respectively without STaSy. Regarding STaSy, we use the reimplementation\footnote{\url{https://github.com/amazon-science/TabSyn/tree/main/baselines/stasy}} provided by Zhang et al.~\yrcite{TabSyn-zhang2023mixedtype}. If the official implementation has code for searching hyperparameters, it is also executed. We employ Optuna~\cite{10.1145/3292500.3330701} for tuning the hyperparameters.

\subsection{Hyperparameter Search Spaces}
We present the hyperparameter search spaces in Tables~\ref{table:trbd-tuning} and \ref{table:others-tuning}. For evaluating sampling quality, we employ RandomForest~\cite{random-forest} and XGBoost~\cite{10.1145/2939672.2939785} as additional classifiers alongside CatBoost. We utilize Optuna for tuning hyperparameters in CatBoost and grid search to identify the optimal hyperparameters for RandomForest and XGBoost.

\begin{table}[h]
  \caption{TRBD hyperparameter spaces from Kotelnikov et al.~\yrcite{tabddpm-pmlr-v202-kotelnikov23a}. Only number of tuning trials is changed to 30.}
  \label{table:trbd-tuning}
  \vskip 0.15in
  \centering
  \begin{tabular}{cc}
    \toprule
    Parameters & Distribution \\
    \midrule
    Learning rate & LogUniform[0.00001, 0.003] \\
    Batch size & Cat\{256, 4096\} \\
    Diffusion timesteps & Cat\{100, 1000\} \\
    Training iterations & Cat\{5000, 10000, 20000\} \\
    \# MLP layers & Int\{2, 4, 6, 8\} \\
    Width of MLP layers & Int\{128, 256, 512, 1024\} \\
    Proportion of samples & Float\{0.25, 0.5, 1, 2, 4, 8\} \\
    \midrule
    Number of tuning trials & 30 \\
    \bottomrule
  \end{tabular}
  \vskip 0.1in
\end{table}

\begin{table}[h]
  \caption{Hyperparameter space of classifier models from Kim et al.~\yrcite{stasy-kim2023}, Lee et al.\yrcite{codi-pmlr-v202-lee23i}, and Kotelnikov et al.~\yrcite{tabddpm-pmlr-v202-kotelnikov23a}}
  \label{table:others-tuning}
  \vskip 0.15in
  \centering
  \begin{tabular}{ccc}
    \toprule
    Models & Parameters & Distribution \\
    \midrule
    \multirow{5}{*}{CatBoost}
    & Learning rate & LogUniform[0.001, 1] \\
    & Depth & UniformInt[3, 10] \\
    & L2 leaf reg & Uniform[0.1, 10.0] \\
    & Bagging temperature & Uniform[0, 1] \\
    & Leaf estimation iterations & UniformInt[1, 10] \\
    \midrule
    \multirow{5}{*}{XGBoost}
    & n\_estimators & Int\{10, 50, 100\} \\
    & min\_child\_weight & Int\{1, 10\} \\
    & max\_depth & Int\{5, 10\} \\
    & gamma & Float\{0.0, 1.0\} \\
    & nthreads & -1 \\
    \midrule
    \multirow{4}{*}{RandomForest}
    & max\_depth & Int\{8, 16, Inf\} \\
    & min\_samples\_split & Int\{2, 4\} \\
    & min\_samples\_leaf & Int\{1, 3\} \\
    & n\_jobs & -1 \\
    \bottomrule
  \end{tabular}
  \vskip 0.1in
\end{table}

\subsection{Datasets}
\label{appendix:datasets}
Table~\ref{table:datasets} shows the information of the 10 datasets used in our experiments. This table does not contain the information of label column. All datasets include both numerical and categorical columns. Eight of these datasets are used in the experiment of TabDDPM~\cite{tabddpm-pmlr-v202-kotelnikov23a}. In addition, we add the two datasets which have high cardinality categorical features.

\begin{table*}[!ht]
\caption{Details of the datasets used in our experiments. \texttt{C} and \texttt{R} denote classification and regression, respectively.}
\label{table:datasets}
\vskip 0.15in
\begin{center}
\begin{small}
\begin{tabular}{lccccccccc}
\toprule
Dataset Name & Abbreviation & \#Num & \#Cat & Task & \#Train & \#Test & \#Val & Cardinalities \\
\midrule
Credit Card (IBM) & CC & 5 & 3 & \texttt{C} & 384,000 & 120,000 & 96,000 & (3, 7515, 153) \\
Airlines & AR & 3 & 4 & \texttt{C} & 345,204 & 107,877 & 86,302 & (18, 6571, 292, 293) \\
Insurance & IS & 3 & 3 & \texttt{R} & 856 & 268 & 214 & (2, 2, 4) \\
Buddy & BD & 4 & 5 & \texttt{C} & 12,053 & 3,767 & 3,014 & (3, 56, 19, 10, 4) \\
Adult & AD & 6 & 8 & \texttt{C} & 26,048 & 16,281 & 6,513 & (9, 16, 7, 15, 6, 5, 2, 42) \\
Churn Modeling & CH & 7 & 4 & \texttt{R} & 6,400 & 2,000 & 1,600 & (3, 2, 2, 2) \\
Cardio & CA & 5 & 6 & \texttt{C} & 44,800 & 11,200 & 14,000 & (2, 3, 3, 2, 2, 2) \\
King & KI & 17 & 3 & \texttt{R} & 13,382 & 4,323 & 3,458 & (2, 5, 5) \\
Abalone & AB & 7 & 1 & \texttt{R} & 2,672 & 836 & 669 & (3) \\
Facebook Comm. Vol. & FB & 36 & 18 & \texttt{R} & 157,638 & 19,720 & 19,722 & (81, others are 2) \\
\bottomrule
\end{tabular}
\end{small}
\end{center}
\vskip 0.1in
\end{table*}

\subsection{Preprocess Datasets}
We preprocess the datasets listed in Table~\ref{table:datasets}. The datasets, excluding CC and AR, are obtained from the official TabDDPM implementation\footnote{\url{https://www.dropbox.com/s/rpckvcs3vx7j605/data.tar?dl=0}}. Consequently, we provide details on the preprocessing of the CC and AR datasets.

\begin{itemize}
  \item \textbf{CC}: We treat \texttt{Use Chip}, \texttt{Merchant State}, and \texttt{Merchant City} as categorical columns, and missing values as one additional category for them. We delete \texttt{User}, \texttt{Errors?}, and \texttt{Merchant Name} columns. We treat other columns as numerical columns and fill missing values with the column’s average. We convert \texttt{Year}, \texttt{Month}, \texttt{Day}, and \texttt{Time} columns into \texttt{Timestamp}. We sample randomly 600k rows.
  \item \textbf{AR}: We treat \texttt{Airline}, \texttt{Flight}, \texttt{AirportFrom}, and \texttt{AirportTo} as categorical columns and other columns as numerical columns. We use all rows.
\end{itemize}

\subsection{Dataset Sources}
\begin{itemize}
  \item \textbf{Credit Card (IBM)}: \url{https://ibm.ent.box.com/v/tabformer-data}
  \item \textbf{Airlines}: \url{https://www.openml.org/search?type=data&sort=runs&id=1169&status=active}
  \item \textbf{Insurance}: \url{https://www.kaggle.com/datasets/mirichoi0218/insurance}
  \item \textbf{Buddy}: \url{https://www.kaggle.com/datasets/akash14/adopt-a-buddy}
  \item \textbf{Churn Modeling}: \url{https://www.kaggle.com/datasets/shrutimechlearn/churn-modelling}
  \item \textbf{Adult}~\cite{10.5555/3001460.3001502}: \url{https://archive.ics.uci.edu/dataset/2/adult} 
  \item \textbf{Cardio}: \url{https://www.kaggle.com/datasets/sulianova/cardiovascular-disease-dataset}
  \item \textbf{King}: \url{https://www.kaggle.com/datasets/harlfoxem/housesalesprediction}
  \item \textbf{Abalone}: \url{https://www.openml.org/search?type=data&sort=runs&id=183&status=active}
  \item \textbf{Facebook Comments}~\cite{10.5555/2867552.2868264}: \url{https://archive.ics.uci.edu/dataset/363/facebook+comment+volume+dataset} 
\end{itemize}

\subsection{Implementation of ResBit}
\label{appendix:resbit-impl}
We present the implementation details of ResBit. Algorithm~\ref{former} provides pseudocode for computing $a_k$ in Equation~\ref{eq:resbit-dim}. The variable \texttt{kind\_of\_cat} represents the number of classes. Algorithm~\ref{latter} is pseudocode for obtaining ResBit for an integer $N$ in the range $[0, M-1]$, where there are $M$ categories.

\begin{minipage}[t]{0.49\linewidth}
\begin{algorithm}[H]
\small
\caption{\small Get $b_1,b_2\ldots$ Algorithm.}
\label{former}
\definecolor{codeblue}{rgb}{0.25,0.5,0.5}
\definecolor{codekw}{rgb}{0.85, 0.18, 0.50}
\lstset{
  backgroundcolor=\color{white},
  basicstyle=\fontsize{7.2pt}{7.2pt}\ttfamily\selectfont,
  columns=fullflexible,
  breaklines=true,
  captionpos=b,
  commentstyle=\fontsize{7.2pt}{7.2pt}\color{codeblue},
  keywordstyle=\fontsize{7.2pt}{7.2pt}\color{codekw},
  escapechar={|}, 
}
\begin{lstlisting}[language=python]
def get_length_ResBit(kind_of_cat: int):
    # get length of ResBit for unique category size kind_of_cat
    res = []
    
    # 0-index
    max_num = kind_of_cat - 1

    while True:
      bits = bin(max_num)[2:]
      
      # in case of M == 2^m
      if "0" not in bits:
          res.append(len(bits))
          break
      
      else:
          s = "1" * (len(bits) - 1)
          max_num -= int(s, base=2)
          res.append(len(s))
          
    return res
\end{lstlisting}
\end{algorithm}
\end{minipage}
\hfill
\begin{minipage}[t]{0.49\linewidth}
\begin{algorithm}[H]
\small
\caption{\small Get ResBit Algorithm for $N$.}
\label{latter}
\definecolor{codeblue}{rgb}{0.25,0.5,0.5}
\definecolor{codekw}{rgb}{0.85, 0.18, 0.50}
\lstset{
  backgroundcolor=\color{white},
  basicstyle=\fontsize{7.5pt}{7.5pt}\ttfamily\selectfont,
  columns=fullflexible,
  breaklines=true,
  captionpos=b,
  commentstyle=\fontsize{7.5pt}{7.5pt}\color{codeblue},
  keywordstyle=\fontsize{7.5pt}{7.5pt}\color{codekw},
  escapechar={|}, 
}
\begin{lstlisting}[language=python]
def int_to_ResBit(N, l):
    # l is returns of algorithm 1
    res = []
    for i in range(len(l)):
        if N == 0:
            res += [0 for _ in range(l[i])]
            continue
        
        bits = "1" * l[i]
        X = int(bits, base=2)
        if X <= N:
            res += [int(z) for z in bits]
            N -= X
            
        else:
            bits_N = bin(N)[2:]
            bits_N = "0" * (l[i] - len(bits_N)) + bits_N
            res += [int(z) for z in bits_N]
            N = 0
            
    return res
\end{lstlisting}
\end{algorithm}
\end{minipage}

\section{Details of the cardinalities survey}
\label{appendix:cardinality}
Table~\ref{table:datasets-comparison} presents our investigation into the cardinality of the categorical features for datasets used in recent tabular data generation research. This table shows the total cardinalities for each data. Only datasets containing both categorical and numerical columns are shown here. Many studies focus on the number of columns and dataset size, with less emphasis on the cardinality of the categorical features. In addition, some datasets have unknown cardinality when preprocessing details are not provided in official implementations, and such cases are marked as \texttt{unknown}.

\begin{table*}[!ht]
\caption{Details of the datasets used in recent researches. \texttt{C} and \texttt{R} denote classification and regression, respectively.}
\label{table:datasets-comparison}
\vskip 0.15in
\begin{center}
\begin{small}
\begin{tabular}{lcccccc}
\toprule
Dataset Name & \#Num & \#Cat & Task & Cardinalities & References \\
\midrule
\multirow{2}{*}{Adult} & 6 & 8 & \multirow{2}{*}{\texttt{C}} & (9,16,7,15,6,5,2,42) & Kotelnikov et al.~\yrcite{tabddpm-pmlr-v202-kotelnikov23a} \\
& 6 & 9 &  & (9,16,7,15,6,5,2,42,2) & Zhang et al.~\yrcite{TabSyn-zhang2023mixedtype} \\
\midrule
\multirow{2}{*}{Default} & 14 & 11 & \multirow{2}{*}{\texttt{C}} & \multirow{2}{*}{(81, 2, 7, 56, 11, 11, 11, 11, 10, 10, 2)} & Zhang et al.~\yrcite{TabSyn-zhang2023mixedtype} \\
& 13 & 11 & & & Kim et al.~\yrcite{stasy-kim2023} \\
\midrule
\multirow{2}{*}{News} & 46 & 2 & \multirow{2}{*}{\texttt{R}} & (6, 1454) & Zhang et al.~\yrcite{TabSyn-zhang2023mixedtype} \\
& 45 & 14 & & \texttt{unknown} & Kim et al.~\yrcite{stasy-kim2023} \\
\midrule
\multirow{2}{*}{Beijing} & 7 & 5 & \multirow{2}{*}{\texttt{R}} & (5, 12, 31, 24, 4) & Zhang et al.~\yrcite{TabSyn-zhang2023mixedtype} \\
& 8 & 6 & & \texttt{unknown} & Kim et al.~\yrcite{stasy-kim2023} \\
\midrule
\multirow{2}{*}{Insurance} & 3 & 3 & \multirow{2}{*}{\texttt{R}} & (2, 2, 4) & Kotelnikov et al.~\yrcite{tabddpm-pmlr-v202-kotelnikov23a} \\
& 3 & 8 & & (49, 2, 2, 2, 2, 2, 2, 4) & Lee et al.~\yrcite{codi-pmlr-v202-lee23i} \\
\midrule
\multirow{2}{*}{Obesity} & 7 & 10 & \multirow{2}{*}{\texttt{C}} & (2, 2, 2, 4, 2, 2, 4, 5, 7, \texttt{unknown}) & Kim et al.~\yrcite{stasy-kim2023} \\ 
& 8 & 9 & & (2, 2, 2, 4, 2, 2, 4, 5, 7) & Lee et al.~\yrcite{codi-pmlr-v202-lee23i} \\
\midrule
Credit & 29 & 1 & \texttt{C} & (2) & \multirow{7}{*}{Kim et al.~\yrcite{stasy-kim2023}} \\
HTRU & 8 & 1 & \texttt{C} & (2) & \\
Spambase & 57 & 1 & \texttt{C} & (2) & \\
Bean & 16 & 1 & \texttt{C} & (7) & \\
Crowdsource & 28 & 1 & \texttt{C} & (6) & \\
Robot & 24 & 1 & \texttt{C} & (4) & \\
Shuttle & 9 & 1 & \texttt{C} & (7) & \\
\midrule
Shoppers & 10 & 8 & \texttt{C} & (10, 8, 13, 9, 20, 3, 2, 2) & \multirow{2}{*}{Zhang et al.~\yrcite{TabSyn-zhang2023mixedtype}, Kim et al.~\yrcite{stasy-kim2023}} \\
Magic & 10 & 1 & \texttt{C} & (2) & \\
\midrule
Buddy & 4 & 5 & \texttt{C} & (3, 56, 19, 10, 4) & \multirow{3}{*}{Kotelnikov et al.~\yrcite{tabddpm-pmlr-v202-kotelnikov23a}} \\
Cardio & 5 & 6 & \texttt{C} & (2, 3, 3, 2, 2, 2) & \\
Churn Modeling & 7 & 4 & \texttt{C} & (3, 2, 2, 2) & \\
\midrule
Bank & 7 & 10 & \texttt{C} & (12, 3, 4, 2, 2, 2, 3, 12, 4, 2) & \multirow{9}{*}{Lee et al.~\yrcite{codi-pmlr-v202-lee23i}} \\
Heart & 4 & 10 & \texttt{C} & (41, 2, 4, 2, 3, 2, 3, 5, 3, 2) & \\
Seismic & 11 & 5 & \texttt{C} & \texttt{unknown} & \\
Stroke & 2 & 8 & \texttt{C} & (3, 2, 2, 2, 5, 2, 4, 2) &  \\
CMC & 2 & 8 & \texttt{C} & (4, 4, 2, 2, 4, 4, 2, 3) & \\
Customer & 5 & 7 & \texttt{C} & (3, 2, 5, 2, 2, 8, 4) & \\
Faults & 24 & 4 & \texttt{C} & (7, \texttt{unknown}) & \\
Absent & 12 & 9 & \texttt{R} & (28, 5, 2, 4, 2, 2, others are unknown) &  \\
Drug & 5 & 1 & \texttt{R} & (3) & \\
\midrule
Abalone & 7 & 1 & \texttt{R} & (3) & \multirow{3}{*}{Kotelnikov et al.~\yrcite{tabddpm-pmlr-v202-kotelnikov23a}} \\
Facebook Comm. Vol. & 36 & 18 & \texttt{R} & (81, others are 2) & \\
King & 17 & 3 & \texttt{R} & (2, 5, 5) & \\
\bottomrule
\end{tabular}
\end{small}
\end{center}
\vskip 0.1in
\end{table*}

\section{Generation Collapse in TabDDPM using Simple Synthetic Data}
\label{appendix:toy-data}
We investigate mode collapse phenomenon in TabDDPM using simple synthetic data. We make data by fixing the numerical and target variables and changing the cardinalities of the categorical variable. The process of the way to make data is shown below.

\begin{enumerate}
  \item Set $N=10000$, the number of rows.
  \item For a numerical column, we set it to normal distribution.
  \item For a target column, we set it to normal distribution.
  \item For a categorical column, we use long tail distribution and uniform distribution respectively.
\end{enumerate}

\begin{figure*}[!htbp]
  \centering
    \begin{tabular}{cc}
    Uniform distribution & Long tail distribution \\
    \midrule
    \\ \multicolumn{2}{c}{50 cardinalities} \\ \\
    \begin{minipage}{0.49\linewidth}
      \centering
      \includegraphics[width=0.95\linewidth]{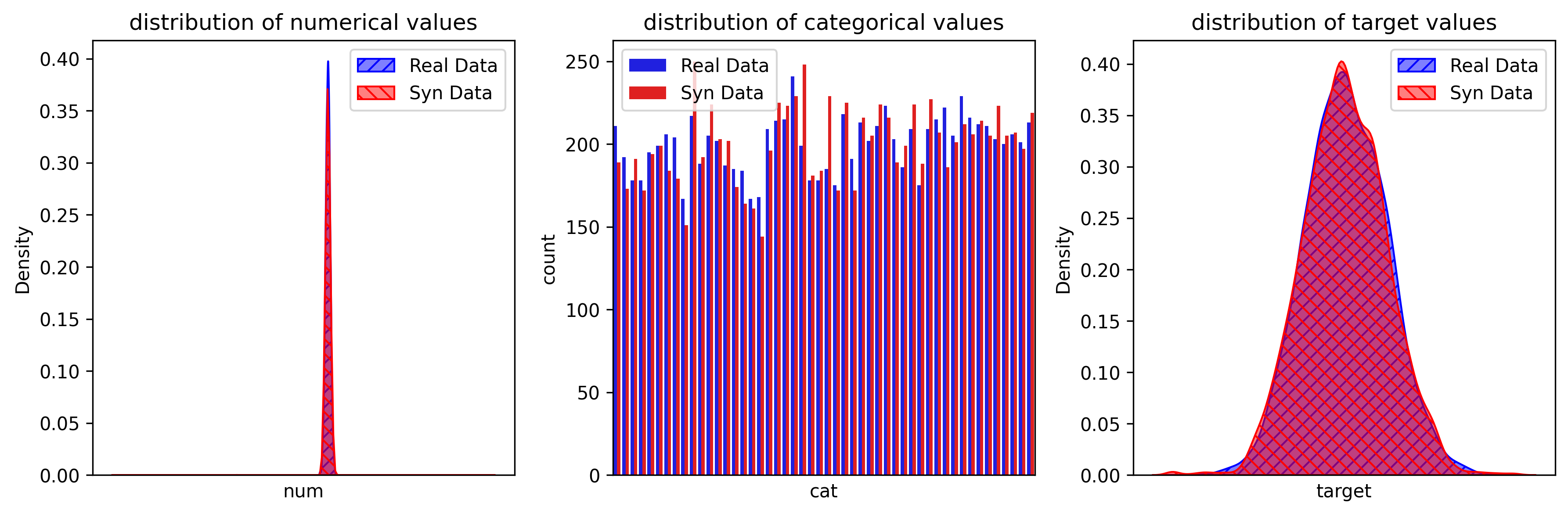}
    \end{minipage} &
    \begin{minipage}{0.49\linewidth}
      \centering
      \includegraphics[width=0.95\linewidth]{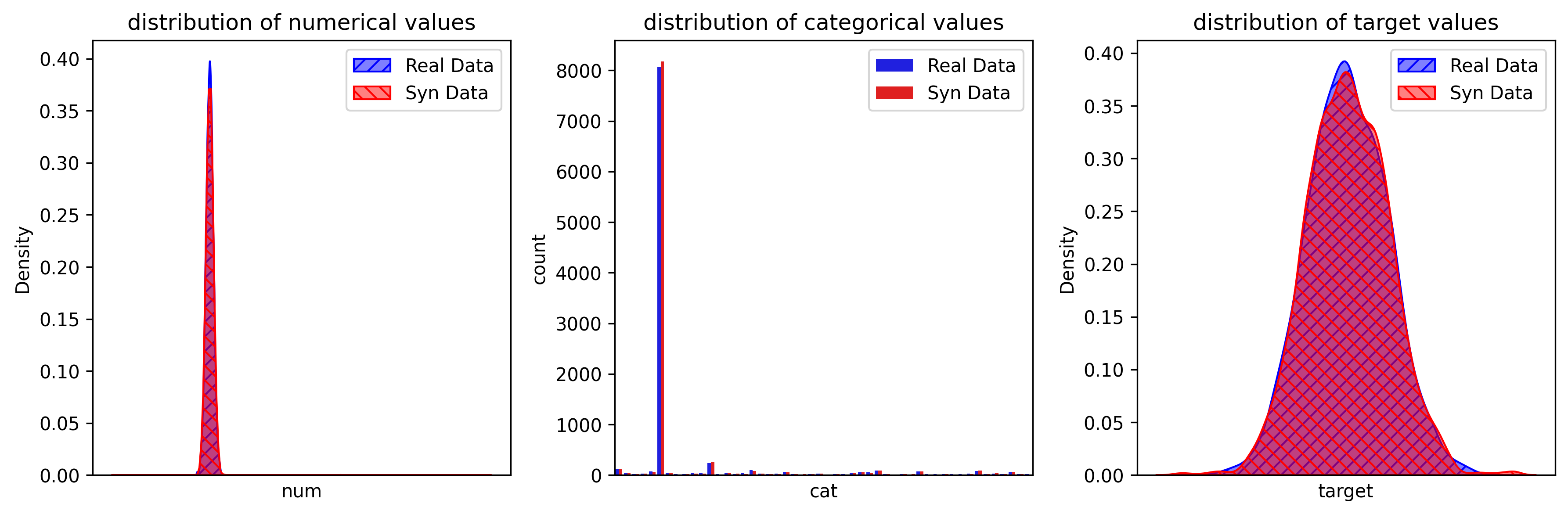}
    \end{minipage}
    \\ \\ \multicolumn{2}{c}{100 cardinalities} \\ \\
    \begin{minipage}{0.49\linewidth}
      \centering
      \includegraphics[width=0.95\linewidth]{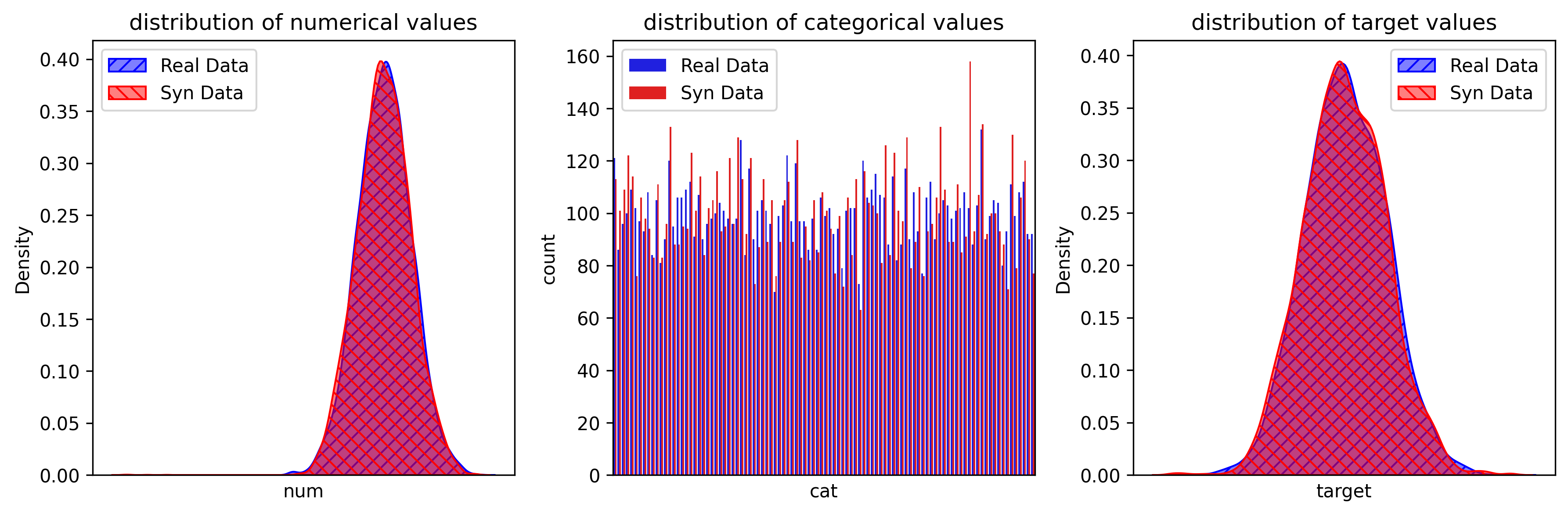}
    \end{minipage} &
    \begin{minipage}{0.49\linewidth}
      \centering
      \includegraphics[width=0.95\linewidth]{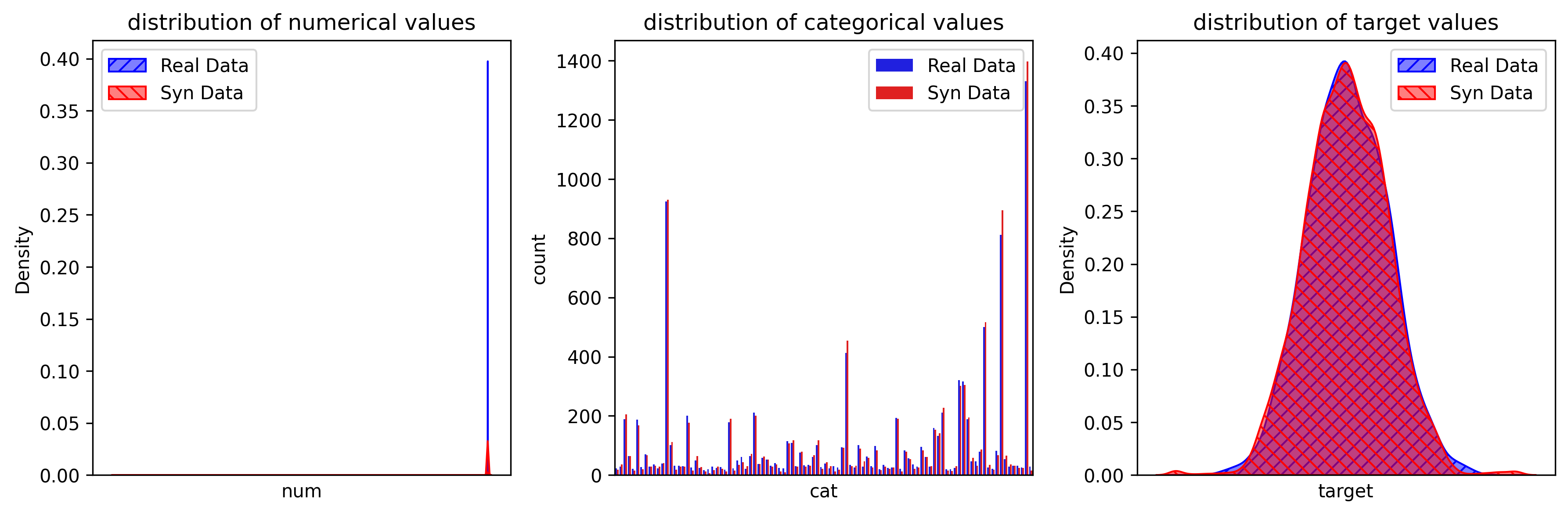}
    \end{minipage}
    \\ \\ \multicolumn{2}{c}{2000 cardinalities} \\ \\
    \begin{minipage}{0.49\linewidth}
      \centering
      \includegraphics[width=0.95\linewidth]{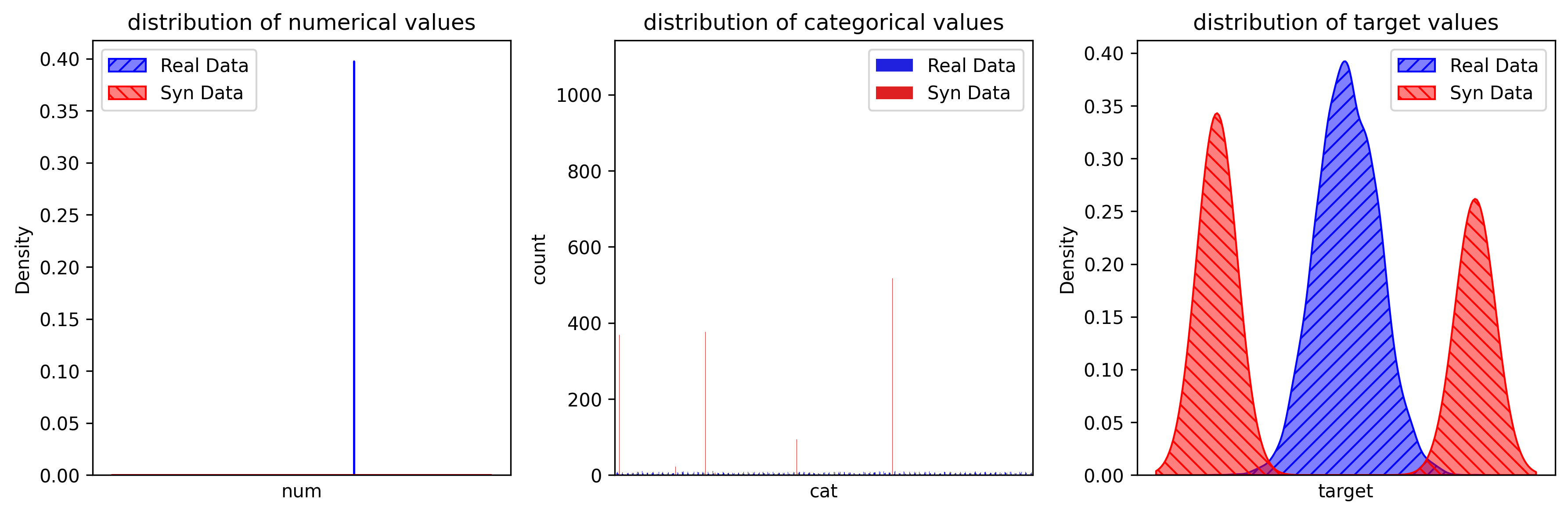}
    \end{minipage} &
    \begin{minipage}{0.49\linewidth}
      \centering
      \includegraphics[width=0.95\linewidth]{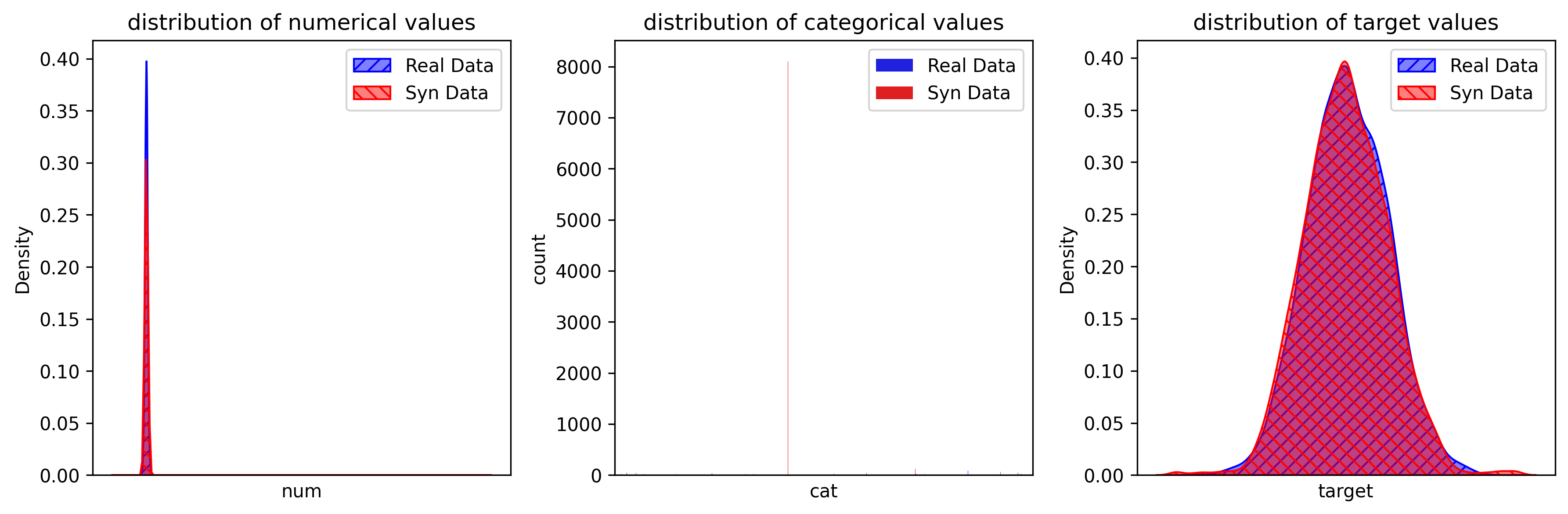}
    \end{minipage}
    \\ \\ \multicolumn{2}{c}{5000 cardinalities} \\ \\
    \begin{minipage}{0.49\linewidth}
      \centering
      \includegraphics[width=0.95\linewidth]{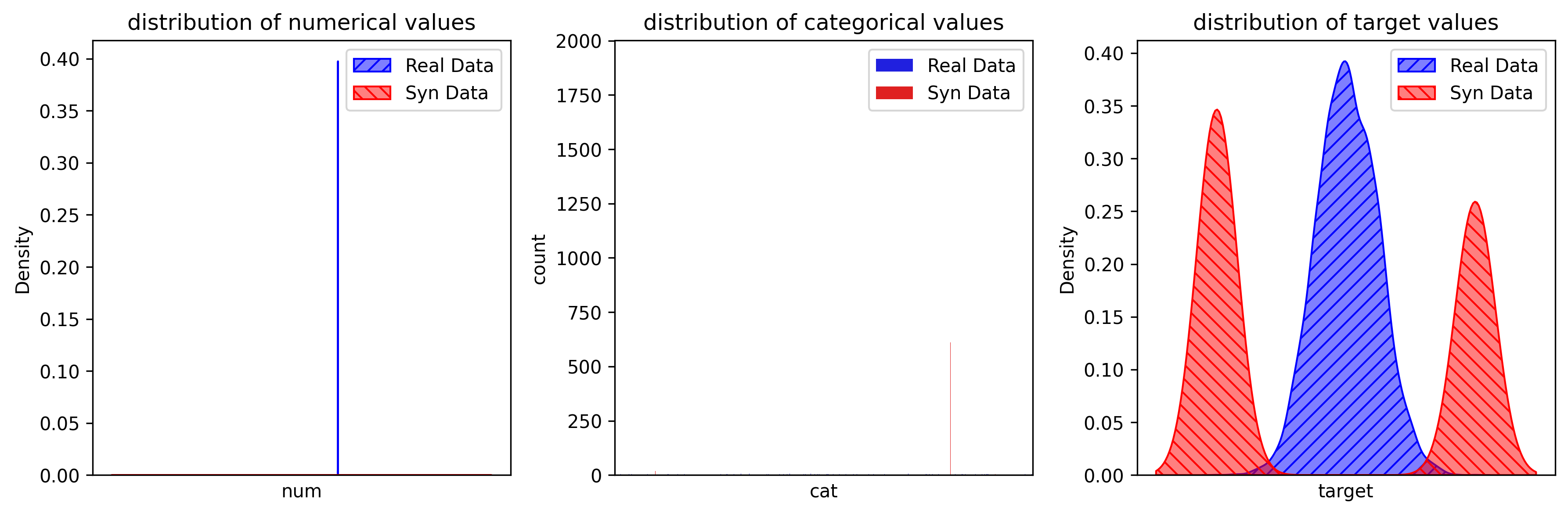}
    \end{minipage} &
    \begin{minipage}{0.49\linewidth}
      \centering
      \includegraphics[width=0.95\linewidth]{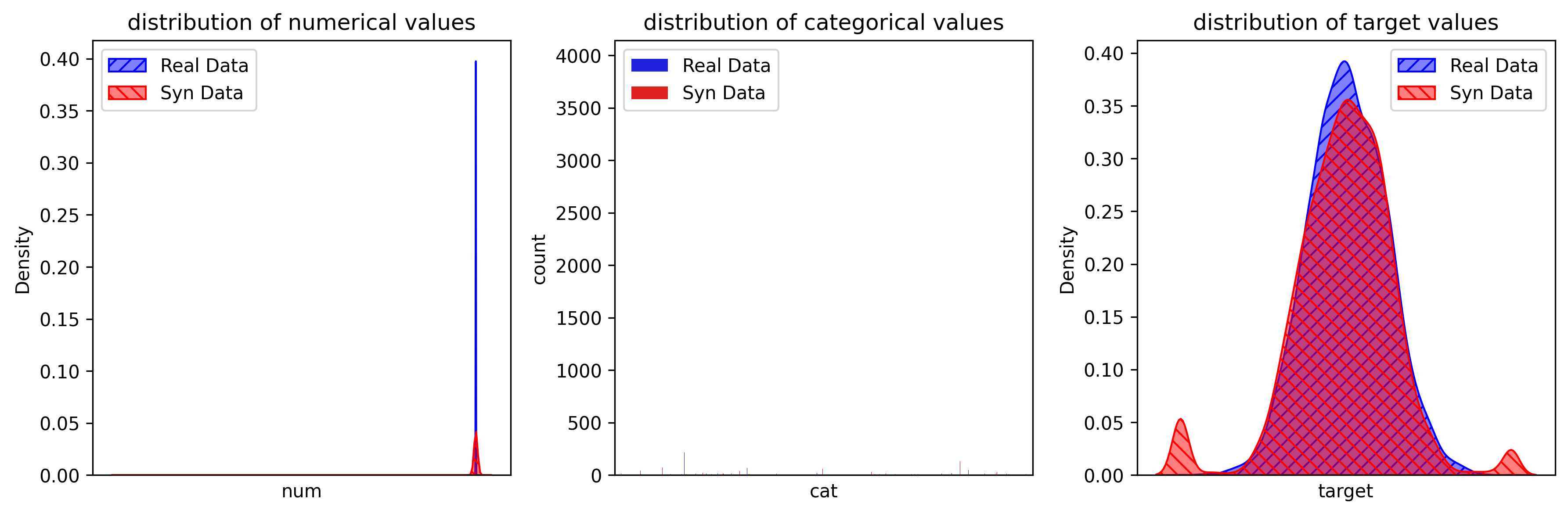}
    \end{minipage}
    \end{tabular}
    \caption{Visualizations of original and generated distributions.}
    \label{fig:comp-dist}
\end{figure*}

The cardinalities we use are 50, 100, 2000, and 5000. Therefore, the number of data types is eight. We train TabDDPM using the data and generate 10000 samples. Figure~\ref{fig:comp-dist} shows the comparison of the distributions between original and generated data each cardinality. We observed only specific categorical values have appeared as the cardinalities increased regardless of what the categorical distribution is. 

\begin{figure*}[!htbp]
  \centering
    \begin{tabular}{cc}
    Uniform distribution & Long tail distribution \\
    \begin{minipage}{0.49\linewidth}
      \centering
      \includegraphics[width=0.95\linewidth]{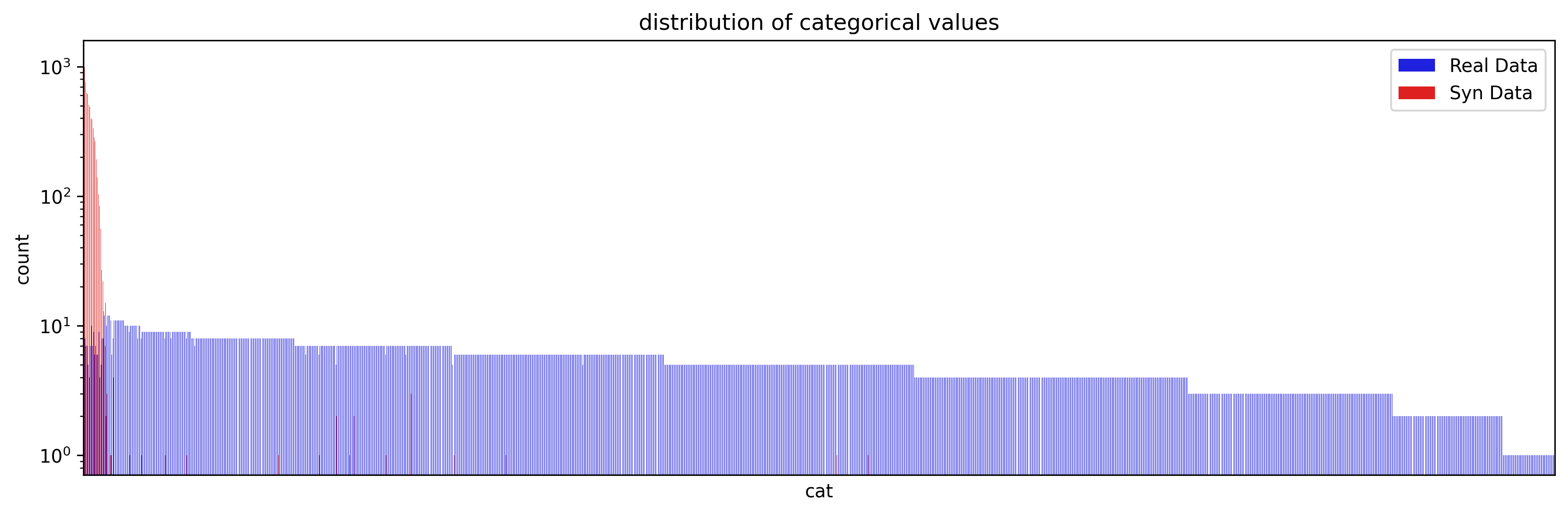}
    \end{minipage} &
    \begin{minipage}{0.49\linewidth}
      \centering
      \includegraphics[width=0.95\linewidth]{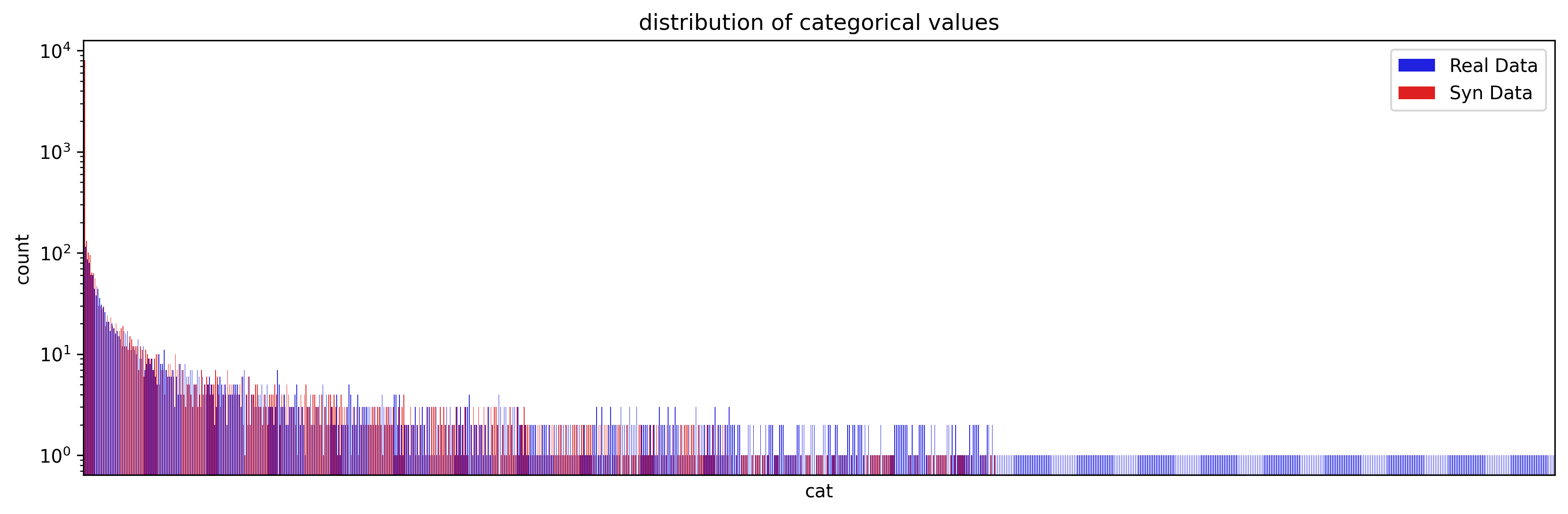}
    \end{minipage}
    \end{tabular}
    \caption{Visualization of original and generated categorical distributions in log scale when the number of cardinality is 2000.}
    \label{fig:log-dist}
\end{figure*}

Moreover, we observe the categorical distributions in case the cardinality is 2000. We show the log scale visualized distributions to recognize the shape of distributions in Figure~\ref{fig:log-dist}. When the categorical distribution is uniform, the model is not learning the distribution at all. On the other hand, when the categorical distribution is long tail, the generated distribution is similar that of original. This is because even if the model generates only certain classes, the shape of the distribution will be like a long tail. Therefore, it is possible to roughly learn the distribution if only the distribution of the top few classes in order of frequency can be learned. This is suggested by the fact that, although the tail part is not generated, the distribution of the top classes appears similar in Figure~\ref{fig:log-dist}. However, when the categorical distribution is uniform, it is necessary for all classes to be generated equally. We consider that it is difficult to learn even the rough shape of the distribution due to this limitation.

\section{Additional Results of Tabular Data Generation}
\label{appendix:additional-results}
\subsection{TSTR Framework}
Tables~\ref{table:cat-f1} and \ref{table:cat-rmse} present the result of TSTR framework using F1 score and RMSE as metrics. The F1 Score registering as 0 in the CC dataset is likely due to the dataset's inherent imbalance. While high cardinalities could contribute, it would be expected for the AR dataset to display a similar behavior if that were the sole cause. However, the results in Table~\ref{table:cat-f1} suggest otherwise. The AR dataset has a balanced $1:1$ dataset, while the CC dataset is highly imbalanced with a ratio of $0.001:0.999$. These observations lead to the inference that extreme imbalance poses a challenge for generative models, resulting in performance degradation.

\begin{table*}[!ht]
\caption{Additional results of our experiments on the classification datasets. We use CatBoost as the classifier.}
\label{table:cat-f1}
\vskip 0.15in
\begin{center}
\begin{tabular}{lcc|cccc}
\toprule
& CC & AR & BD & AD & CH & CA \\
\midrule
Real     
& .684{\tiny$\pm$.067} 
& .587{\tiny$\pm$.004} 
& .987{\tiny$\pm$.000} 
& .716{\tiny$\pm$.003} 
& .580{\tiny$\pm$.033} 
& .738{\tiny$\pm$.001} \\
\midrule
TVAE
& \texttt{Failed}
& .508{\tiny$\pm$.002}
& .846{\tiny$\pm$.006}
& .622{\tiny$\pm$.008}
& .470{\tiny$\pm$.020}
& .510{\tiny$\pm$.019} \\
CTGAN
& \texttt{Failed}
& \textbf{.584{\tiny$\pm$.012}}
& .820{\tiny$\pm$.001}
& .527{\tiny$\pm$.007}
& .249{\tiny$\pm$.029}
& .701{\tiny$\pm$.007} \\
TabDDPM  
& .000{\tiny$\pm$.000} 
& .279{\tiny$\pm$.134} 
& \textbf{.907{\tiny$\pm$.003}}
& .673{\tiny$\pm$.002} 
& .601{\tiny$\pm$.007} 
& \textbf{.737{\tiny$\pm$.001}} \\
CoDi     
& \texttt{N/A} 
& \texttt{N/A} 
& .286{\tiny$\pm$.030} 
& .357{\tiny$\pm$.046} 
& .023{\tiny$\pm$.010} 
& .656{\tiny$\pm$.008} \\
STaSy    
& \texttt{N/A} 
& \texttt{N/A} 
& .843{\tiny$\pm$.010} 
& .623{\tiny$\pm$.008} 
& .574{\tiny$\pm$.013} 
& .729{\tiny$\pm$.003} \\
TabSyn   
& \textbf{.293{\tiny$\pm$.083}}
& .557{\tiny$\pm$.004}
& .905{\tiny$\pm$.008} 
& .639{\tiny$\pm$.009} 
& .594{\tiny$\pm$.010} 
& .721{\tiny$\pm$.002} \\
\midrule
TRBD (ours) 
& .000{\tiny$\pm$.000} 
& .482{\tiny$\pm$.010} 
& \textbf{.901{\tiny$\pm$.007}}
& \textbf{.676{\tiny$\pm$.004}}
& \textbf{.612{\tiny$\pm$.009}}
& \textbf{.737{\tiny$\pm$.002}} \\
\bottomrule
\end{tabular}
\end{center}
\vskip 0.1in
\end{table*}

\begin{table*}[!ht]
\caption{Additional results of our experiments on the regression datasets. We use CatBoost as the classifier.}
\label{table:cat-rmse}
\vskip 0.15in
\begin{center}
\begin{tabular}{lcccc}
\toprule
& IS & KI & AB & FB \\
\midrule
Real     
& 4,475.287{\tiny$\pm$390.028} 
& 107,633.528{\tiny$\pm$1,022.334} 
& 2.009{\tiny$\pm$.004} 
& 5.297{\tiny$\pm$.019} \\
\midrule
TVAE
& 7,769.982{\tiny$\pm$270.354}
& 230,264.317{\tiny$\pm$505.661}
& 2.581{\tiny$\pm$.011}
& 9.115{\tiny$\pm$.056} \\
CTGAN
& 16,786.700{\tiny$\pm$467.101}
& 237,360.868{\tiny$\pm$3,803.299}
& 2.556{\tiny$\pm$.052}
& 10.224{\tiny$\pm$.102} \\
TabDDPM  
& 4,050.754{\tiny$\pm$47.229} 
& 129,491.185{\tiny$\pm$5,426.840} 
& \textbf{1.986{\tiny$\pm$.028}}
& 7.648{\tiny$\pm$.084} \\
CoDi     
& 28,657.977{\tiny$\pm$1,643.204} 
& 3,461,545.525{\tiny$\pm$412,674.821} 
& 6.177{\tiny$\pm$.373} 
& \texttt{N/A} \\
STaSy    
& 7,930.060{\tiny$\pm$857.998} 
& 141,939.588{\tiny$\pm$5,090.508} 
& 3.857{\tiny$\pm$.057} 
& 7.710{\tiny$\pm$.133} \\
TabSyn   
& 5,097.570{\tiny$\pm$59.577} 
& 141,801.203{\tiny$\pm$2,817.380} 
& 2.101{\tiny$\pm$.012} 
& 8.309{\tiny$\pm$.094} \\
\midrule
TRBD (ours)  
& \textbf{3,785.253{\tiny$\pm$69.238}}
& \textbf{115,411.676{\tiny$\pm$1,972.408}}
& \textbf{2.002{\tiny$\pm$.013}}
& \textbf{7.127{\tiny$\pm$.057}}  \\
\bottomrule
\end{tabular}
\end{center}
\vskip 0.1in
\end{table*}

Tables~\ref{table:xgb-roc} to \ref{table:xgb-rmse} present the results using XGBoost as the classifier, while Tables~\ref{table:rf-auroc} to \ref{table:rf-rmse} show the results using RandomForest are shown. Similar trends are observed across other classifiers, as seen in the case of CatBoost.

\begin{table*}[!ht]
\caption{Results of our experiments on the classification datasets. We use XGBoost as the classifier and AUROC as the metric.}
\label{table:xgb-roc}
\vskip 0.15in
\begin{center}
\begin{tabular}{lcc|cccc}
\toprule
& CC & AR & BD & AD & CH & CA \\
\midrule
Real     
& .887{\tiny$\pm$.185} 
& .721{\tiny$\pm$.001} 
& .987{\tiny$\pm$.001} 
& .916{\tiny$\pm$.002} 
& .868{\tiny$\pm$.007} 
& .799{\tiny$\pm$.003} \\
\midrule
TVAE
& .500{\tiny$\pm$.000}
& .638{\tiny$\pm$.001}
& .980{\tiny$\pm$.001}
& .869{\tiny$\pm$.002}
& .760{\tiny$\pm$.006}
& .645{\tiny$\pm$.007} \\
CTGAN
& .500{\tiny$\pm$.000}
& .647{\tiny$\pm$.003}
& .962{\tiny$\pm$.003}
& .877{\tiny$\pm$.003}
& .683{\tiny$\pm$.009}
& .758{\tiny$\pm$.003} \\
TabDDPM  
& .429{\tiny$\pm$.138}
& .514{\tiny$\pm$.007}
& .935{\tiny$\pm$.007}
& \textbf{.902{\tiny$\pm$.001}}
& .890{\tiny$\pm$.008}
& .797{\tiny$\pm$.001} \\
CoDi     
& \texttt{N/A} 
& \texttt{N/A} 
& .525{\tiny$\pm$.021} 
& .657{\tiny$\pm$.049} 
& .625{\tiny$\pm$.013} 
& .577{\tiny$\pm$.054} \\
STaSy    
& \texttt{N/A} 
& \texttt{N/A} 
& .980{\tiny$\pm$.001} 
& .897{\tiny$\pm$.002} 
& .839{\tiny$\pm$.006} 
& .791{\tiny$\pm$.001} \\
TabSyn   
& \textbf{.916{\tiny$\pm$.011}}
& \textbf{.690{\tiny$\pm$.001}}
& \textbf{.984{\tiny$\pm$.001}}
& .895{\tiny$\pm$.002} 
& .839{\tiny$\pm$.002} 
& .789{\tiny$\pm$.001} \\
\midrule
TRBD (ours) 
& .643{\tiny$\pm$.014}
& .619{\tiny$\pm$.002}
& .922{\tiny$\pm$.002}
& \textbf{.901{\tiny$\pm$.001}}
& \textbf{.977{\tiny$\pm$.003}}
& \textbf{.801{\tiny$\pm$.000}} \\
\bottomrule
\end{tabular}
\end{center}
\vskip 0.1in
\end{table*}

\begin{table*}[!ht]
\caption{Results of our experiments on the classification datasets. We use XGBoost as the classifier and F1 score as the metric.}
\label{table:xgb-f1}
\vskip 0.15in
\begin{center}
\begin{tabular}{lcc|cccc}
\toprule
& CC & AR & BD & AD & CH & CA \\
\midrule
Real     
& .620{\tiny$\pm$.066} 
& .595{\tiny$\pm$.002} 
& .927{\tiny$\pm$.005} 
& .689{\tiny$\pm$.004} 
& .585{\tiny$\pm$.018} 
& .719{\tiny$\pm$.004} \\
\midrule
TVAE
& .000{\tiny$\pm$.000}
& .525{\tiny$\pm$.003}
& .850{\tiny$\pm$.012}
& .620{\tiny$\pm$.008}
& .467{\tiny$\pm$.014}
& .519{\tiny$\pm$.017} \\
CTGAN
& .000{\tiny$\pm$.000}
& \textbf{.590{\tiny$\pm$.005}}
& .826{\tiny$\pm$.002}
& .571{\tiny$\pm$.030}
& .252{\tiny$\pm$.026}
& .707{\tiny$\pm$.005} \\
TabDDPM  
& .000{\tiny$\pm$.000}
& .353{\tiny$\pm$.095}
& .573{\tiny$\pm$.002}
& \textbf{.661{\tiny$\pm$.003}}
& .669{\tiny$\pm$.013} 
& \textbf{.724{\tiny$\pm$.004}} \\
CoDi     
& \texttt{N/A} 
& \texttt{N/A} 
& .333{\tiny$\pm$.016} 
& .409{\tiny$\pm$.033} 
& .154{\tiny$\pm$.081} 
& .650{\tiny$\pm$.015} \\
STaSy    
& \texttt{N/A} 
& \texttt{N/A} 
& .841{\tiny$\pm$.015} 
& .614{\tiny$\pm$.018} 
& .592{\tiny$\pm$.013} 
& \textbf{.727{\tiny$\pm$.005}} \\
TabSyn   
& \textbf{.172{\tiny$\pm$.066}}
& .554{\tiny$\pm$.003}
& \textbf{.905{\tiny$\pm$.005}}
& .629{\tiny$\pm$.014} 
& .569{\tiny$\pm$.002} 
& .719{\tiny$\pm$.003} \\
\midrule
TRBD (ours) 
& .000{\tiny$\pm$.000}
& .485{\tiny$\pm$.004}
& .568{\tiny$\pm$.002}
& \textbf{.663{\tiny$\pm$.002}}
& \textbf{.854{\tiny$\pm$.011}}
& \textbf{.724{\tiny$\pm$.001}} \\
\bottomrule
\end{tabular}
\end{center}
\vskip 0.1in
\end{table*}

\begin{table*}[!ht]
\caption{Results of our experiments on the regression datasets. We use XGBoost as the classifier and $R^2$ score as the metric.}
\label{table:xgb-r2}
\vskip 0.15in
\begin{center}
\begin{tabular}{lcccc}
\toprule
& IS & KI & AB & FB \\
\midrule
Real     
& .858{\tiny$\pm$.021} 
& .885{\tiny$\pm$.003} 
& .537{\tiny$\pm$.042} 
& .791{\tiny$\pm$.007} \\
\midrule
TVAE
& .665{\tiny$\pm$.049}
& .641{\tiny$\pm$.006}
& .286{\tiny$\pm$.007}
& .541{\tiny$\pm$.043} \\
CTGAN
& $-0.577${\tiny$\pm$.044}
& .580{\tiny$\pm$.009}
& .288{\tiny$\pm$.032}
& .420{\tiny$\pm$.032} \\
TabDDPM  
& .903{\tiny$\pm$.008} 
& .817{\tiny$\pm$.007} 
& .553{\tiny$\pm$.004} 
& .615{\tiny$\pm$.001} \\
CoDi     
& $-3.766${\tiny$\pm$.393} 
& $-64.883${\tiny$\pm$14.632} 
& $-1.583${\tiny$\pm$.322} 
& \texttt{N/A} \\
STaSy    
& .744{\tiny$\pm$.039} 
& .848{\tiny$\pm$.011} 
& $-0.201${\tiny$\pm$.075} 
& .678{\tiny$\pm$.016} \\
TabSyn   
& .843{\tiny$\pm$.004} 
& .835{\tiny$\pm$.011} 
& .515{\tiny$\pm$.006} 
& .638{\tiny$\pm$.006} \\
\midrule
TRBD (ours) 
& \textbf{.915{\tiny$\pm$.003}}
& \textbf{.900{\tiny$\pm$.010}}
& \textbf{.568{\tiny$\pm$.006}}
& \textbf{.689{\tiny$\pm$.006}} \\
\bottomrule
\end{tabular}
\end{center}
\vskip 0.1in
\end{table*}

\begin{table*}[!ht]
\caption{Results of our experiments on the regression datasets. We use XGBoost as the classifier and RMSE as the metric.}
\label{table:xgb-rmse}
\vskip 0.15in
\begin{center}
\begin{tabular}{lcccc}
\toprule
& IS & KI & AB & FB \\
\midrule
Real     
& 4,777.171{\tiny$\pm$690.880} 
& 129,298.734{\tiny$\pm$8,184.660} 
& 2.301{\tiny$\pm$.120} 
& 6.051{\tiny$\pm$.098} \\
\midrule
TVAE
& 7,423.982{\tiny$\pm$535.919}
& 233,179.555{\tiny$\pm$2,057.680}
& 2.566{\tiny$\pm$.013}
& 9.084{\tiny$\pm$.042} \\
CTGAN
& 16,140.365{\tiny$\pm$223.186}
& 252,229.633{\tiny$\pm$2,844.116}
& 2.562{\tiny$\pm$.057}
& 10.405{\tiny$\pm$.279} \\
TabDDPM  
& 3,872.860{\tiny$\pm$156.814} 
& 166,357.334{\tiny$\pm$2,973.649} 
& 2.031{\tiny$\pm$.009} 
& 8.323{\tiny$\pm$.135} \\
CoDi     
& 26,698.977{\tiny$\pm$1,107.887} 
& 2,978,663.320{\tiny$\pm$340,768.191} 
& 5.292{\tiny$\pm$.346} 
& \texttt{N/A} \\
STaSy    
& 6,488.204{\tiny$\pm$499.168} 
& 151,618.842{\tiny$\pm$5,597.809} 
& 3.327{\tiny$\pm$.103} 
& 7.612{\tiny$\pm$.191} \\
TabSyn   
& 5,091.418{\tiny$\pm$59.761} 
& 158,079.866{\tiny$\pm$5,064.771} 
& 2.115{\tiny$\pm$.013} 
& 8.079{\tiny$\pm$.067} \\
\midrule
TRBD (ours) 
& \textbf{3,630.228{\tiny$\pm$69.255}}
& \textbf{123,291.742{\tiny$\pm$5,934.827}}
& \textbf{1.995{\tiny$\pm$.015}}
& \textbf{7.486{\tiny$\pm$.069}} \\
\bottomrule
\end{tabular}
\end{center}
\vskip 0.1in
\end{table*}

\begin{table*}[!ht]
\caption{Results of our experiments on the classification datasets. We use RandomForest as the classifier and AUROC as the metric.}
\label{table:rf-auroc}
\vskip 0.15in
\begin{center}
\begin{tabular}{lcc|cccc}
\toprule
& CC & AR & BD & AD & CH & CA \\
\midrule
Real     
& .948{\tiny$\pm$.011} 
& .725{\tiny$\pm$.001} 
& .986{\tiny$\pm$.002} 
& .926{\tiny$\pm$.002} 
& .859{\tiny$\pm$.004} 
& .798{\tiny$\pm$.004} \\
\midrule
TVAE
& \texttt{Failed}
& .633{\tiny$\pm$.001}
& .980{\tiny$\pm$.000}
& .876{\tiny$\pm$.002}
& .763{\tiny$\pm$.007}
& .718{\tiny$\pm$.004} \\
CTGAN
& \texttt{Failed}
& .648{\tiny$\pm$.002}
& .965{\tiny$\pm$.002}
& .883{\tiny$\pm$.002}
& .680{\tiny$\pm$.009}
& .766{\tiny$\pm$.002} \\
TabDDPM  
& .511{\tiny$\pm$.054} 
& .499{\tiny$\pm$.018}
& .978{\tiny$\pm$.002}
& \textbf{.906{\tiny$\pm$.001}}
& .895{\tiny$\pm$.005}
& .799{\tiny$\pm$.001} \\
CoDi     
& \texttt{N/A} 
& \texttt{N/A} 
& .551{\tiny$\pm$.040} 
& .601{\tiny$\pm$.039} 
& .630{\tiny$\pm$.016} 
& .562{\tiny$\pm$.030} \\
STaSy    
& \texttt{N/A} 
& \texttt{N/A} 
& .980{\tiny$\pm$.001} 
& .898{\tiny$\pm$.001} 
& .836{\tiny$\pm$.007} 
& .794{\tiny$\pm$.002} \\
TabSyn   
& \textbf{.913{\tiny$\pm$.007}}
& \textbf{.687{\tiny$\pm$.001}}
& \textbf{.984{\tiny$\pm$.001}}
& .895{\tiny$\pm$.002} 
& .838{\tiny$\pm$.002} 
& .794{\tiny$\pm$.001} \\
\midrule
TRBD (ours) 
& .615{\tiny$\pm$.054}
& .637{\tiny$\pm$.002}
& .975{\tiny$\pm$.002}
& .903{\tiny$\pm$.001}
& \textbf{.970{\tiny$\pm$.003}}
& \textbf{.804{\tiny$\pm$.001}} \\
\bottomrule
\end{tabular}
\end{center}
\vskip 0.1in
\end{table*}

\begin{table*}[!ht]
\caption{Results of our experiments on the classification datasets. We use RandomForest as the classifier and F1 as the metric.}
\label{table:rf-f1}
\vskip 0.15in
\begin{center}
\begin{tabular}{lcc|cccc}
\toprule
& CC & AR & BD & AD & CH & CA \\
\midrule
Real     
& .597{\tiny$\pm$.040}
& .572{\tiny$\pm$.003}
& .930{\tiny$\pm$.004}
& .689{\tiny$\pm$.004} 
& .527{\tiny$\pm$.034}
& .716{\tiny$\pm$.006} \\
\midrule
TVAE
& \texttt{Failed}
& .496{\tiny$\pm$.005}
& .838{\tiny$\pm$.001}
& .604{\tiny$\pm$.013}
& .468{\tiny$\pm$.030}
& .532{\tiny$\pm$.017} \\
CTGAN
& \texttt{Failed}
& \textbf{.589{\tiny$\pm$.003}}
& .821{\tiny$\pm$.005}
& .571{\tiny$\pm$.030}
& .252{\tiny$\pm$.026}
& .713{\tiny$\pm$.002} \\
TabDDPM  
& .000{\tiny$\pm$.000}
& .136{\tiny$\pm$.087}
& .790{\tiny$\pm$.013}
& \textbf{.661{\tiny$\pm$.007}}
& .671{\tiny$\pm$.007}
& .717{\tiny$\pm$.005} \\
CoDi     
& \texttt{N/A} 
& \texttt{N/A} 
& .321{\tiny$\pm$.054} 
& .395{\tiny$\pm$.009} 
& .067{\tiny$\pm$.029} 
& .630{\tiny$\pm$.049} \\
STaSy    
& \texttt{N/A} 
& \texttt{N/A} 
& .821{\tiny$\pm$.008} 
& .599{\tiny$\pm$.011} 
& .583{\tiny$\pm$.021} 
& .728{\tiny$\pm$.004} \\
TabSyn   
& \textbf{.041{\tiny$\pm$.027}}
& .547{\tiny$\pm$.001}
& \textbf{.895{\tiny$\pm$.006}}
& .599{\tiny$\pm$.019} 
& .565{\tiny$\pm$.021} 
& .717{\tiny$\pm$.003} \\
\midrule
TRBD (ours) 
& .000{\tiny$\pm$.000}
& .479{\tiny$\pm$.002}
& .751{\tiny$\pm$.050}
& \textbf{.661{\tiny$\pm$.003}}
& \textbf{.864{\tiny$\pm$.007}}
& \textbf{.728{\tiny$\pm$.002}} \\
\bottomrule
\end{tabular}
\end{center}
\vskip 0.1in
\end{table*}

\begin{table*}[!ht]
\caption{Results of our experiments on the regression datasets. We use RandomForest as the classifier and $R^2$ score as the metric.}
\label{table:rf-r2}
\vskip 0.15in
\begin{center}
\begin{tabular}{lcccc}
\toprule
& IS & KI & AB & FB \\
\midrule
Real     
& .831{\tiny$\pm$.052}
& .868{\tiny$\pm$.009} 
& .558{\tiny$\pm$.032} 
& .790{\tiny$\pm$.007} \\
\midrule
TVAE
& .693{\tiny$\pm$.031}
& .636{\tiny$\pm$.010}
& .308{\tiny$\pm$.006}
& .567{\tiny$\pm$.056} \\
CTGAN
& $-0.618${\tiny$\pm$.048}
& .569{\tiny$\pm$.014}
& .281{\tiny$\pm$.032}
& .436{\tiny$\pm$.012} \\
TabDDPM  
& .903{\tiny$\pm$.006}
& .817{\tiny$\pm$.022} 
& .547{\tiny$\pm$.011} 
& .653{\tiny$\pm$.010} \\
CoDi     
& $-4.100${\tiny$\pm$.365}  
& $-81.550${\tiny$\pm$10.043} 
& $-2.019${\tiny$\pm$.417} 
& \texttt{N/A} \\
STaSy    
& .772{\tiny$\pm$.018} 
& .820{\tiny$\pm$.011} 
& $-0.500${\tiny$\pm$.046} 
& .681{\tiny$\pm$.013} \\
TabSyn   
& .861{\tiny$\pm$.034} 
& .825{\tiny$\pm$.009} 
& .515{\tiny$\pm$.007} 
& .656{\tiny$\pm$.008} \\
\midrule
TRBD (ours) 
& \textbf{.913{\tiny$\pm$.003}}
& \textbf{.899{\tiny$\pm$.014}}
& \textbf{.559{\tiny$\pm$.001}}
& \textbf{.695{\tiny$\pm$.004}}  \\
\bottomrule
\end{tabular}
\end{center}
\vskip 0.1in
\end{table*}

\begin{table*}[!ht]
\caption{Results of our experiments on the regression datasets. We use RandomForest as the classifier and RMSE as the metric.}
\label{table:rf-rmse}
\vskip 0.15in
\begin{center}
\begin{tabular}{lcccc}
\toprule
& IS & KI & AB & FB \\
\midrule
Real     
& 4,737.314{\tiny$\pm$330.580}
& 137,629.411{\tiny$\pm$7,872.392} 
& 2.106{\tiny$\pm$.121} 
& 6.183{\tiny$\pm$.141} \\
\midrule
TVAE
& 7,115.766{\tiny$\pm$356.529}
& 234,758.355{\tiny$\pm$3,234.940}
& 2.527{\tiny$\pm$.011}
& 8.819{\tiny$\pm$.555} \\
CTGAN
& 16,349.869{\tiny$\pm$241.497}
& 255,570.930{\tiny$\pm$4,097.423}
& 2.575{\tiny$\pm$.058}
& 10.086{\tiny$\pm$.109} \\
TabDDPM  
& 3,880.795{\tiny$\pm$115.701}
& 166,210.395{\tiny$\pm$9,867.884} 
& 2.044{\tiny$\pm$.024} 
& 7.907{\tiny$\pm$.111} \\
CoDi     
& 27,625.513{\tiny$\pm$1,000.384} 
& 3,349,602.925{\tiny$\pm$206,738.563} 
& 5.718{\tiny$\pm$.406} 
& \texttt{N/A} \\
STaSy    
& 6,129.863{\tiny$\pm$236.914} 
& 164,822.726{\tiny$\pm$4,872.698} 
& 3.720{\tiny$\pm$.058} 
& 7.581{\tiny$\pm$.153} \\
TabSyn   
& 5,083.900{\tiny$\pm$72.430} 
& 162,895.708{\tiny$\pm$4,128.800} 
& 2.115{\tiny$\pm$.014} 
& 7.876{\tiny$\pm$.095} \\
\midrule
TRBD (ours) 
& \textbf{3,676.466{\tiny$\pm$66.751}}
& \textbf{131,291.742{\tiny$\pm$7,757.758}}
& \textbf{2.017{\tiny$\pm$.022}}
& \textbf{7.411{\tiny$\pm$.054}} \\
\bottomrule
\end{tabular}
\end{center}
\vskip 0.1in
\end{table*}

\subsection{Sample-wise Quality Score of Generated Data}
We evaluate the generated data using statistical metrics. It is essential to note that these results should not be evaluated in isolation. As demonstrated in Section~\ref{subsec:exp-cardinality}, accurate reproduction of categorical data does not inherently guarantee high quality. In other words, statistical excellence does not necessarily imply high quality. In generative tasks, the generated data is often used to train new machine learning models, and the primary evaluation should focus on downstream tasks. Hence, we consider statistical metrics as supplementary evaluations. It is not advisable to solely determine the superiority or inferiority of models based on these results. Following Liu et al.~\yrcite{liu2023goggle} and Zhang et al.~\yrcite{TabSyn-zhang2023mixedtype}, we use $\alpha$-Precision and $\beta$-Recall proposed in Alaa et al.~\yrcite{pmlr-v162-alaa22a}. $\alpha$-Precision evaluates how faithfully the generated data adheres to the distribution of real data. $\beta$-Recall assesses the coverage of the generated data. 

Tables~\ref{table:alpha-precision} and \ref{table:beta-recall} present the result of the two metrics. TVAE and CTGAN achieve high scores in $\alpha$-Precision for the CC dataset, indicating a considerable fidelity to real data. However, as shown in Table~\ref{table:tstr-result-auroc}, the generated data fails to reproduce the imbalance. This emphasizes that statistical metrics do not guarantee sampling quality.

In Sections~\ref{subsec:tstr-exp}, \ref{subsec:exp-cardinality}, and \ref{subsec:exp-runtime}, TabSyn and TRBD demonstrated competitive results. This trend is also reflected in Table~\ref{table:alpha-precision}. However, as shown in Table~\ref{table:beta-recall}, TRBD outperforms TabSyn in $\beta$-Recall. TabSyn transforms data into latent variables, potentially causing the loss of rare information during this process. Since the generative model cannot discern the importance of the lost rare information, the performance of TabSyn may degrade in more extensive experiments.

\begin{table*}[!ht]
\caption{Comparison of $\alpha$-Precision scores. Higher value indicates better score.}
\label{table:alpha-precision}
\vskip 0.15in
\begin{center}
\begin{small}
\begin{tabular}{lcc|cccccccc}
\toprule
& CC & AR & IS & BD & AD & CH & CA & KI & AB & FB \\
\midrule
TVAE
& .960{\tiny$\pm$.001}
& .809{\tiny$\pm$.001}
& .901{\tiny$\pm$.009}
& .960{\tiny$\pm$.004}
& .856{\tiny$\pm$.004}
& .547{\tiny$\pm$.005}
& .788{\tiny$\pm$.002} 
& .918{\tiny$\pm$.002}
& .898{\tiny$\pm$.011}
& .008{\tiny$\pm$.002} \\
CTGAN
& .861{\tiny$\pm$.001}
& .915{\tiny$\pm$.001}
& .722{\tiny$\pm$.010}
& .739{\tiny$\pm$.004}
& .717{\tiny$\pm$.006}
& .894{\tiny$\pm$.006}
& .746{\tiny$\pm$.002}
& .465{\tiny$\pm$.002}
& .594{\tiny$\pm$.005}
& \textbf{.934{\tiny$\pm$.002}} \\
TabDDPM
& .047{\tiny$\pm$.000}
& .061{\tiny$\pm$.000}
& \textbf{.974{\tiny$\pm$.010}}
& .973{\tiny$\pm$.008}
& .933{\tiny$\pm$.003}
& .933{\tiny$\pm$.005}
& .975{\tiny$\pm$.004}
& .968{\tiny$\pm$.002}
& .908{\tiny$\pm$.005}
& .578{\tiny$\pm$.006} \\
CoDi     
& \texttt{N/A}
& \texttt{N/A} 
& .368{\tiny$\pm$.012}
& .105{\tiny$\pm$.002}
& .031{\tiny$\pm$.004}
& .387{\tiny$\pm$.004} 
& .070{\tiny$\pm$.023}
& .033{\tiny$\pm$.012}
& .059{\tiny$\pm$.001}
& \texttt{N/A} \\
STaSy    
& \texttt{N/A}
& \texttt{N/A}
& .359{\tiny$\pm$.009}
& .871{\tiny$\pm$.002}
& .638{\tiny$\pm$.007}
& .718{\tiny$\pm$.003}
& .933{\tiny$\pm$.006} 
& .830{\tiny$\pm$.005}
& .829{\tiny$\pm$.008}
& .902{\tiny$\pm$.010} \\
TabSyn   
& \textbf{.977{\tiny$\pm$.001}}
& \textbf{.958{\tiny$\pm$.001}}
& \textbf{.971{\tiny$\pm$.010}}
& .933{\tiny$\pm$.002}
& .980{\tiny$\pm$.004}
& \textbf{.990{\tiny$\pm$.004}}
& \textbf{.988{\tiny$\pm$.002}}
& \textbf{.983{\tiny$\pm$.007}}
& \textbf{.983{\tiny$\pm$.007}}
& .400{\tiny$\pm$.005} \\
\midrule
TRBD (ours) 
& .866{\tiny$\pm$.001}
& .541{\tiny$\pm$.001}
& \textbf{.974{\tiny$\pm$.008}}
& \textbf{.988{\tiny$\pm$.007}}
& \textbf{.988{\tiny$\pm$.003}}
& \textbf{.984{\tiny$\pm$.007}}
& .981{\tiny$\pm$.002} 
& .962{\tiny$\pm$.003}
& \textbf{.983{\tiny$\pm$.006}}
& \textbf{.935{\tiny$\pm$.002}} \\
\bottomrule
\end{tabular}
\end{small}
\end{center}
\vskip 0.1in
\end{table*}

\begin{table*}[!ht]
\caption{Comparison of $\beta$-Recall scores. Higher value indicates better score.}
\label{table:beta-recall}
\vskip 0.15in
\begin{center}
\begin{small}
\begin{tabular}{lcc|cccccccc}
\toprule
& CC & AR & IS & BD & AD & CH & CA & KI & AB & FB \\
\midrule
TVAE
& .167{\tiny$\pm$.001}
& .014{\tiny$\pm$.000}
& .274{\tiny$\pm$.012}
& .201{\tiny$\pm$.003}
& .347{\tiny$\pm$.004}
& .090{\tiny$\pm$.002}
& .265{\tiny$\pm$.006} 
& .194{\tiny$\pm$.006}
& .179{\tiny$\pm$.009}
& .000{\tiny$\pm$.000} \\
CTGAN
& .072{\tiny$\pm$.001}
& .031{\tiny$\pm$.000}
& .159{\tiny$\pm$.005}
& .271{\tiny$\pm$.006}
& .267{\tiny$\pm$.004}
& .232{\tiny$\pm$.004}
& .330{\tiny$\pm$.004}
& .110{\tiny$\pm$.002}
& .008{\tiny$\pm$.005}
& .086{\tiny$\pm$.004} \\
TabDDPM  
& .000{\tiny$\pm$.000}
& .000{\tiny$\pm$.000}
& .694{\tiny$\pm$.008}
& \textbf{.565{\tiny$\pm$.004}}
& \textbf{.503{\tiny$\pm$.005}}
& .506{\tiny$\pm$.002}
& \textbf{.502{\tiny$\pm$.009}}
& .376{\tiny$\pm$.006}
& .458{\tiny$\pm$.013}
& .261{\tiny$\pm$.005} \\
CoDi     
& \texttt{N/A}
& \texttt{N/A} 
& .110{\tiny$\pm$.009}
& .006{\tiny$\pm$.001}
& .002{\tiny$\pm$.000}
& .070{\tiny$\pm$.003} 
& .007{\tiny$\pm$.001}
& .001{\tiny$\pm$.000}
& .002{\tiny$\pm$.001}
& \texttt{N/A} \\
STaSy    
& \texttt{N/A}
& \texttt{N/A}
& .063{\tiny$\pm$.010}
& .326{\tiny$\pm$.005}
& .306{\tiny$\pm$.006}
& .365{\tiny$\pm$.007}
& .471{\tiny$\pm$.008} 
& .350{\tiny$\pm$.005}
& .440{\tiny$\pm$.006}
& \textbf{.394{\tiny$\pm$.005}} \\
TabSyn   
& \textbf{.331{\tiny$\pm$.001}}
& \textbf{.095{\tiny$\pm$.001}}
& .503{\tiny$\pm$.014}
& .500{\tiny$\pm$.005}
& .435{\tiny$\pm$.006}
& .504{\tiny$\pm$.003}
& .488{\tiny$\pm$.007} 
& .381{\tiny$\pm$.006}
& .461{\tiny$\pm$.006}
& .129{\tiny$\pm$.005} \\
\midrule
TRBD (ours) 
& .136{\tiny$\pm$.001}
& .042{\tiny$\pm$.000}
& \textbf{.742{\tiny$\pm$.016}}
& .498{\tiny$\pm$.002}
& .445{\tiny$\pm$.004}
& \textbf{.650{\tiny$\pm$.002}}
& \textbf{.499{\tiny$\pm$.003}}
& \textbf{.548{\tiny$\pm$.002}}
& \textbf{.485{\tiny$\pm$.007}}
& .182{\tiny$\pm$.003} \\
\bottomrule
\end{tabular}
\end{small}
\end{center}
\vskip 0.1in
\end{table*}

\subsection{Runtime}
We present the additional results of the runtime between TabDDPM and TRBD. The batch size for training is 4,096, and the number of iterations is kept the same for TabDDPM and TRBD. However, since the number of layers in the MLPs used in the architectures is different, we also provide an approximate count of the parameters.

Table~\ref{table:training-time-additional} shows the comparison of the training time. From this table, it can be observed that even in cases where the number of parameters is higher, TRBD operates faster compared to TabDDPM, indicating that the computation of multinomial diffusion is the bottleneck. Additionally, on high cardinality datasets, despite having a higher number of parameters, poor results are obtained in terms of diversity and performance in downstream tasks. Therefore, it is unlikely that increasing the number of parameters will lead to a solution to the problem.

\begin{table*}[!ht]
\caption{Comparison of the training time and the number of parameters between TabDDPM and TRBD.}
\label{table:training-time-additional}
\vskip 0.15in
\begin{center}
\begin{small}
\begin{tabular}{llcc|cccccccc}
\toprule
& & CC & AR & IS & BD & AD & CH & CA & KI & AB & FB \\
\midrule
\multirow{2}{*}{Time} & 
TabDDPM 
& 3,494s
& 3,014s
& 273s
& 471s
& 691s
& 567s
& 696s
& 600s
& 60.5s
& 648s \\
& TRBD (ours) 
& \textbf{183s}
& \textbf{265s}
& \textbf{113s}
& \textbf{454s}
& \textbf{495s}
& \textbf{264s}
& \textbf{404s}
& \textbf{234s}
& \textbf{41.9s}
& \textbf{372s} \\
\midrule
\multirow{2}{*}{\#params} & 
TabDDPM 
& 14.8M
& 13.1M
& 1.1M
& 1.3M
& 3.8M
& 1.3M
& 4.3M
& 5.4M
& 2.4M
& 0.7M \\
& TRBD (ours)
& 0.4M
& 2.5M
& 1.5M
& 4.0M
& 4.9M
& 2.7M
& 3.7M
& 2.8M
& 1.6M
& 4.0M \\
\bottomrule
\end{tabular}
\end{small}
\end{center}
\vskip 0.1in
\end{table*}

Table~\ref{table:sampling-time-additional} shows the comparison of the sampling time. We use the same model as during training and generate 300k samples for the CC and AR datasets, and 26k samples for the other datasets. The batch size during generation is set to match the number of samples generated, and the sampling steps are unified to $T=1000$.

It can be observed that similar results to those during training comparison are obtained. However, during generation, there is not a significant reduction in time, possibly because the bottleneck of multinomial diffusion is less prominent. Additionally, on high cardinality datasets, TabDDPM experiences \texttt{OOM} errors. This indicates that memory is exhausted due to the large dimensionality of one-hot vectors.

\begin{table*}[!ht]
\caption{Comparison of the sampling time and the number of parameters between TabDDPM and TRBD.}
\label{table:sampling-time-additional}
\vskip 0.15in
\begin{center}
\begin{small}
\begin{tabular}{llcc|cccccccc}
\toprule
& & CC & AR & IS & BD & AD & CH & CA & KI & AB & FB \\
\midrule
\multirow{2}{*}{Time} & 
TabDDPM 
& \texttt{OOM}
& \texttt{OOM}
& 12.0s
& \textbf{14.0s}
& 25.8s
& 22.8s
& 25.3s
& 14.5s
& 16.2s
& 21.4s \\
& TRBD (ours) 
& \textbf{4.3s}
& \textbf{196s}
& \textbf{9.4s}
& 21.0s
& \textbf{24.3s}
& \textbf{13.6s}
& \textbf{18.2s}
& \textbf{13.7s}
& \textbf{11.3s}
& \textbf{19.1s} \\
\midrule
\multirow{2}{*}{\#params} & 
TabDDPM 
& 14.8M
& 13.1M
& 1.1M
& 1.3M
& 3.8M
& 1.3M
& 4.3M
& 5.4M
& 2.4M
& 0.7M \\
& TRBD (ours)
& 0.4M
& 2.5M
& 1.5M
& 4.0M
& 4.9M
& 2.7M
& 3.7M
& 2.8M
& 1.6M
& 4.0M \\
\bottomrule
\end{tabular}
\end{small}
\end{center}
\vskip 0.1in
\end{table*}

\subsection{ResBit with STaSy}
We present the results of introducing ResBit to STaSy, along with the training times. For the evaluation metrics of TSTR, we use AUROC for the classification datasets and $R^2$ score for the regression datasets. We use CatBoost as the classifier.

Tables~\ref{table:stasy-num} and \ref{table:stasy-cat} present the results. It can be observed that training proceeded faster due to the reduction in input dimensions. Among the low cardinality datasets, AUROC is maintained for the classification datasets, and improvement in $R^{2}$ score is observed for most regression datasets. In high cardinality datasets, it can be confirmed that training completes within 24 hours due to the reduction in dimensions. For example, in the case of the AR dataset, with one-hot vectors, the dimensionality is $3+18+6571+292+293=7177$. However, using ResBit, it is reduced to fewer than $100$ dimensions.

\begin{table*}[!htbp]
\caption{Results of TSTR framework in classification datasets. Metric is AUROC.}
\label{table:stasy-num}
\vskip 0.15in
\begin{center}
\begin{small}
\begin{tabular}{llcc|cccc}
\toprule
& & CC & AR  & BD & AD & CH & CA \\
\midrule
\multirow{2}{*}{TSTR} &
STaSy    
& \texttt{N/A} 
& \texttt{N/A} 
& .978{\tiny$\pm$.002} 
& \textbf{.900{\tiny$\pm$.002}}
& \textbf{.835{\tiny$\pm$.004}}
& \textbf{.791{\tiny$\pm$.001}} \\
& STaSy with ResBit
& \textbf{.785{\tiny$\pm$.022}}
& \textbf{.549{\tiny$\pm$.033}}
& \textbf{.980{\tiny$\pm$.001}}
& .898{\tiny$\pm$.003}
& \textbf{.834{\tiny$\pm$.004}}
& \textbf{.790{\tiny$\pm$.003}} \\
\midrule
\multirow{2}{*}{Training Time} & 
STaSy    
& \texttt{N/A} 
& \texttt{N/A} 
& 4,707s
& 6,676s
& 3,879s
& 9,071s \\
& STaSy with ResBit
& \textbf{57,854s}
& \textbf{52,557s}
& \textbf{3,211s}
& \textbf{5,394s}
& \textbf{2,408s}
& \textbf{7,659s} \\
\bottomrule
\end{tabular}
\end{small}
\end{center}
\vskip 0.1in
\end{table*}

\begin{table*}[!htbp]
\caption{Results of TSTR framework in regression datasets. Metric is $R^2$ score.}
\label{table:stasy-cat}
\vskip 0.15in
\begin{center}
\begin{small}
\begin{tabular}{llccccc}
\toprule
& & IS & KI & AB & FB \\
\midrule
\multirow{2}{*}{TSTR} &
STaSy    
& .618{\tiny$\pm$.080} 
& \textbf{.867{\tiny$\pm$.010}}
& $-0.613${\tiny$\pm$.047} 
& .670{\tiny$\pm$.011} \\
& STaSy with ResBit
& \textbf{.686{\tiny$\pm$.007}}
& \textbf{.867{\tiny$\pm$.007}}
& \textbf{.479{\tiny$\pm$.005}}
& \textbf{.691{\tiny$\pm$.010}} \\
\midrule
\multirow{2}{*}{Training Time} & 
STaSy    
& 2,137s
& 4,937s
& 3,205s
& 26,406s \\
& STaSy with ResBit
& \textbf{1,493s}
& \textbf{3,477s}
& \textbf{1,820s}
& \textbf{24,494s} \\
\bottomrule
\end{tabular}
\end{small}
\end{center}
\vskip 0.1in
\end{table*}

\section{Additional Applications of ResBit}
\label{appendix:others}
We demonstrate the additional applications of ResBit. While it is applicable in scenarios involving one-hot vectors, we specifically explore its utility in image classification and class conditional image generation. It is important to emphasize that this demonstration does not claim an improvement in accuracy.

\subsection{Image Classification}
In the image classification model, we modify the output of the final layer. To ensure a fair comparison, we round output and evaluate its performance based on whether it matches the label representation (one-hot and ResBit). For one-hot vectors, a well-trained model should achieve performance comparable to top-1 accuracy. We use SGD~\cite{pmlr-v28-sutskever13} as the optimizer and Binary Cross Entropy with Sigmoid as the loss function. The dataset we use is Food 101~\cite{10.1007/978-3-319-10599-4_29}. The learning rate is set as shown in Table~\ref{table:larning-rate}, and ResNet34~\cite{7780459} is employed as the classification model, with training conducted for 100 epochs with a batch size of 1,024.

\begin{table}[!ht]
\caption{Learning rate in the training in the image classification.}
\label{table:larning-rate}
\vskip 0.15in
\begin{center}
\begin{tabular}{cc}
\toprule
Epochs (1-index) & Learning Rate \\
\midrule
1-30 & 0.1 \\
31-60 &	0.02 \\
61-80 & 0.004 \\
81-100 & 0.0008 \\
\bottomrule
\end{tabular}
\end{center}
\vskip 0.1in
\end{table}

The results are presented in Table~\ref{table:food-classification}. The \#Classes indicates the number of classes used from the dataset, with a total of 101 classes. While both methods show a decrease in accuracy with an increase in dimensionality, ResBit can mitigate the reduction in accuracy due to its ability to restrain dimensionality growth. However, in the case of one-hot vectors, the accuracy is only slightly higher than random guessing with 101 classes, indicating the difficulty of the task.

\begin{table*}[!ht]
\caption{Results of image classification experiments.}
\label{table:food-classification}
\vskip 0.15in
\begin{center}
\begin{tabular}{cccccc}
\toprule
 & \multicolumn{2}{c}{\#Label dim} & \multicolumn{2}{c}{Accuracy(\%)} \\
\#Classes & one-hot & ResBit & one-hot & ResBit \\
\midrule
10 & 10 & 5 & \textbf{52.32} & 50.00 \\
20 & 20 & 7 & \textbf{34.38} & 34.04 \\
40 & 40 & 9 & 20.47 & \textbf{26.67} \\
60 & 60 & 16 & 9.97 & \textbf{18.15} \\
80 & 80 & 11 & 6.31 & \textbf{24.85} \\
101 & 101 & 15 & 1.61 & \textbf{21.59} \\
\bottomrule
\end{tabular}
\end{center}
\vskip 0.1in
\end{table*}

\subsection{Class Conditional Image Generation}
\begin{figure}[htbp]
  \centering
  \begin{minipage}[b]{0.45\textwidth}
    \centering
    \includegraphics[width=0.8\linewidth]{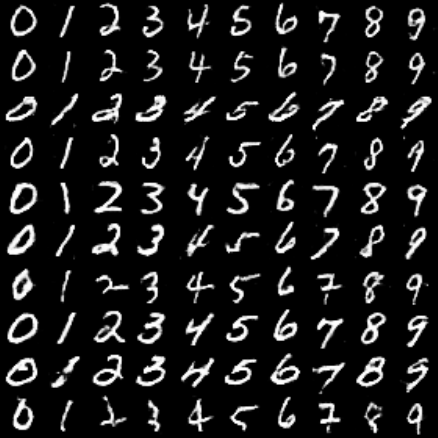}
    \subcaption{One-hot conditioning}
  \end{minipage}
  \begin{minipage}[b]{0.45\textwidth}
    \centering
    \includegraphics[width=0.8\linewidth]{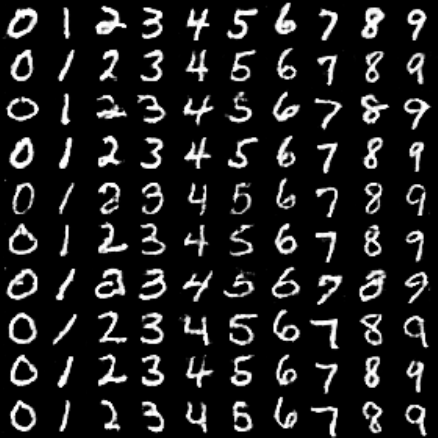}
    \subcaption{ResBit conditioning (ours)}
  \end{minipage}
  \caption{Class conditioned samples from InfoGAN trained on MNIST}
  \label{MNIST-result}
\end{figure}

\begin{figure}[htbp]
  \centering
  \begin{minipage}[b]{0.45\linewidth}
    \centering
    \includegraphics[width=0.8\linewidth]{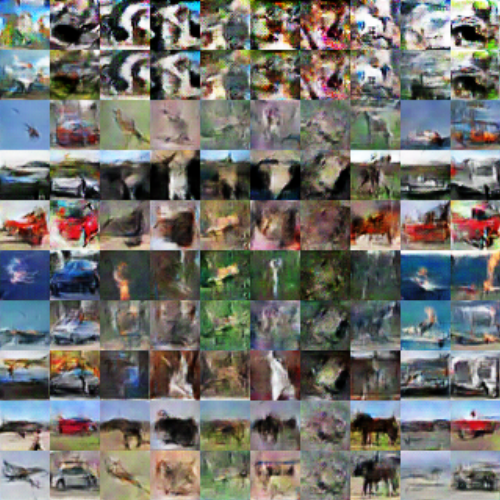}
    \subcaption{One-hot conditioning}
  \end{minipage}
  \begin{minipage}[b]{0.45\linewidth}
    \centering
    \includegraphics[width=0.8\linewidth]{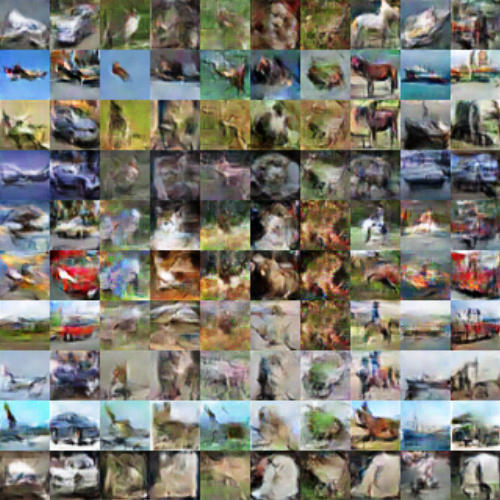}
    \subcaption{ResBit conditioning (ours)}
  \end{minipage}
  \caption{Class conditioned samples from InfoGAN trained on CIFAR-10}
  \label{CIFAR10-vis-result}
\end{figure}

\begin{figure}[htbp]
  \centering
  \begin{minipage}[b]{0.45\linewidth}
    \centering
    \includegraphics[width=0.8\linewidth]{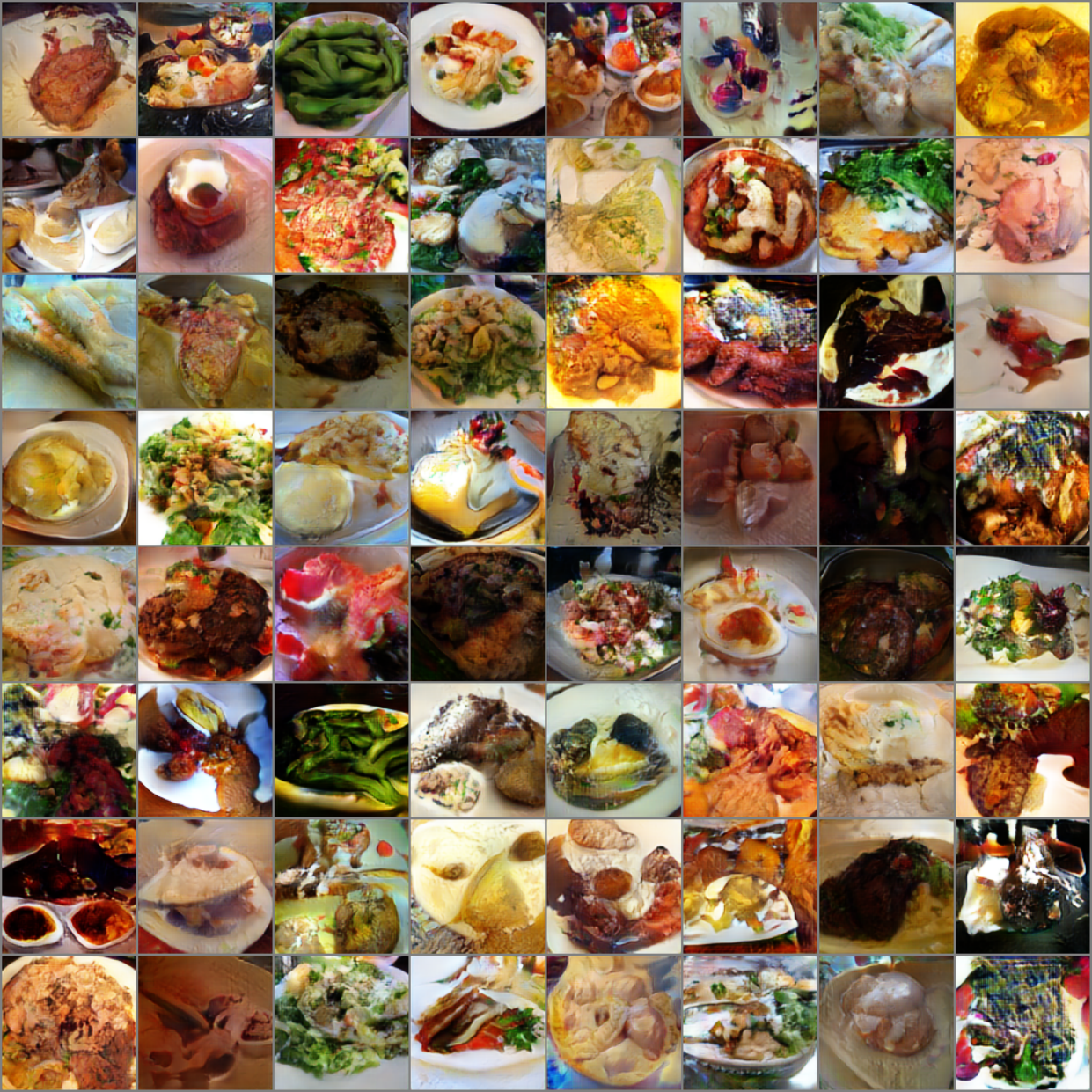}
    \subcaption{One-hot conditioning}
  \end{minipage}
  \begin{minipage}[b]{0.45\linewidth}
    \centering
    \includegraphics[width=0.8\linewidth]{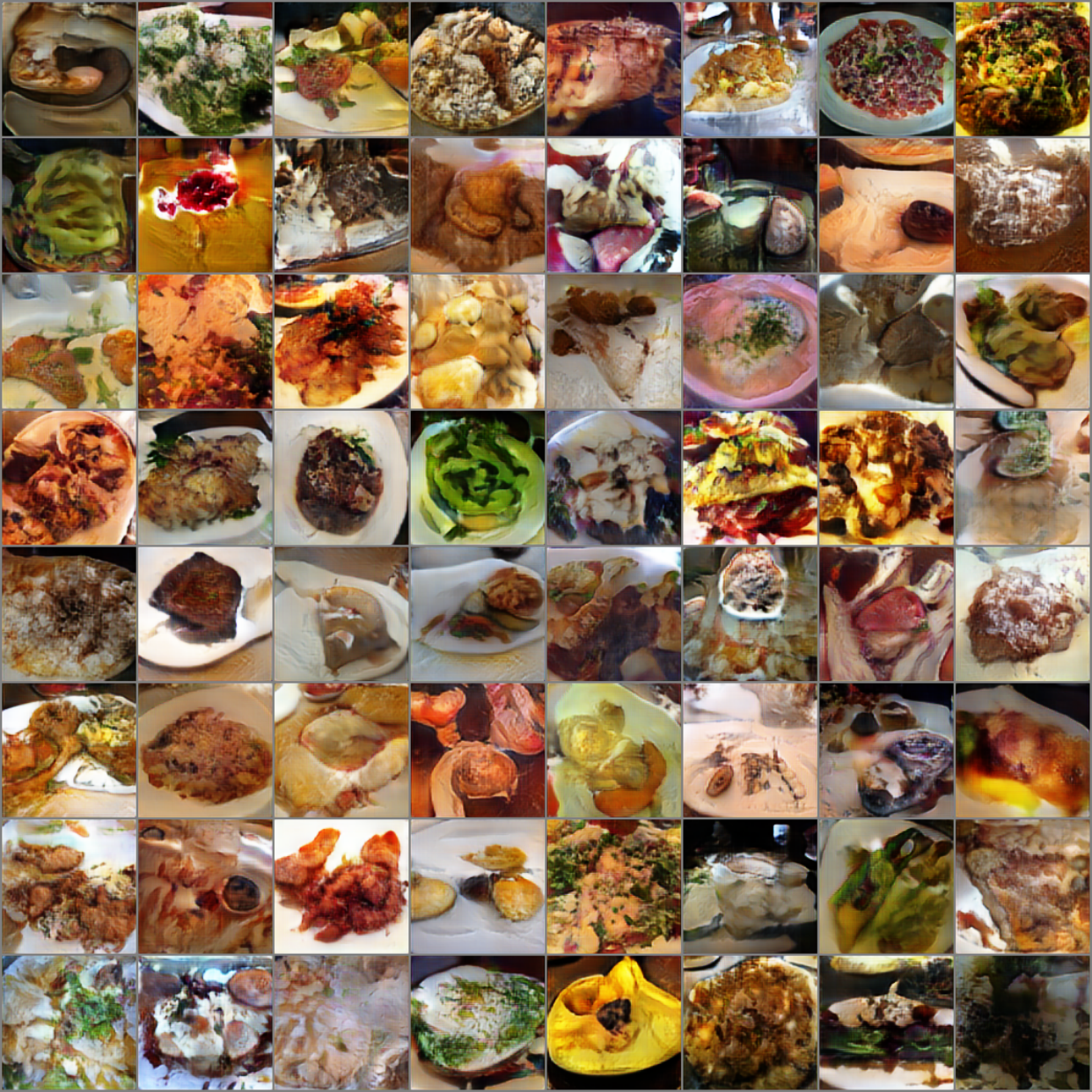}
    \subcaption{ResBit conditioning (ours)}
  \end{minipage}
  \caption{Generated samples from SNGAN trained on Food101}
  \label{food-result}
\end{figure}

We conduct experiments using Conditional Generative Adversarial Networks (CGAN)~\cite{mirza2014conditional} as a method to specify classes. In other words, we concatenate a noise vector and a conditional vector representing the class and input them to the Generator. We use the CIFAR-10~\cite{Krizhevsky09learningmultiple}, MNIST~\cite{lecun2010mnist}, and Food101 datasets. For the architectures, we utilize InfoGAN~\cite{10.5555/3157096.3157340} for CIFAR-10 and MNIST, and SNGAN~\cite{miyato2018spectral} for Food101. We use Adam~\cite{kingma2017adam} as the optimizer with hyperparameters $lr=0.0002, \beta_1=0.5, \beta_2=0.999$. Our primary objective is not to improve the quality of generated images, so we do not incorporate accuracy improvement techniques. However, in the case of SNGAN, we aim for training stability by updating the discriminator five times for every single update of the generator. We conduct training for 30 epochs on MNIST, 300 epochs on CIFAR-10, and updated the generator 15,000 times for Food101. Results are presented in Figures~\ref{MNIST-result}, \ref{CIFAR10-vis-result}, and \ref{food-result}. It is evident that conditioning works for all datasets. In MNIST, ResBit lacks diversity, while in CIFAR-10, visually, it can be observed that one-hot has more mode collapse.

For CIFAR-10, we conduct a comparison using the FID~\cite{NIPS2017_8a1d6947} score.  We calculate the FID score by generating 50,000 images for each method. The results are presented in Table~\ref{table:fid}. It is noticeable that there is no significant decrease in accuracy due to the change in the conditioning method. In addition, with conditioning, the ``out-of-index'' problem does not occur, making the conditioning with binary, which has the fewest dimensions, yield the best results.

In our experiments, we use a small number of classes. This decision is based on the observation that the FID score tends to degrade when the number of classes decreases or the number of images per class decreases in class-conditional GANs~\cite{shahbazi2022collapse}.

\begin{table*}[htbp]
\caption{Comparison of the FID score on CIFAR-10 dataset.}
\label{table:fid}
\vskip 0.15in
\begin{center}
\begin{tabular}{cccc}
\toprule
& one-hot & ResBit (ours) & Binary \\
\midrule
FID & 95.52 & 82.21 & \textbf{63.54} \\
Label dim & 10 & 5 & 4 \\
\bottomrule
\end{tabular}
\end{center}
\vskip 0.1in
\end{table*}

%%%%%%%%%%%%%%%%%%%%%%%%%%%%%%%%%%%%%%%%%%%%%%%%%%%%%%%%%%%%%%%%%%%%%%%%%%%%%%%

\end{document}